\documentclass[a4paper, 12pt]{article}

\usepackage{comment}
\usepackage{natbib}
\usepackage{url}
\setcitestyle{aysep={}} 
\usepackage{setspace}
\usepackage{subcaption}
\usepackage{tikz}
\usetikzlibrary{matrix,chains,positioning,decorations.pathreplacing,arrows}

\usepackage[nottoc]{tocbibind}
\usepackage[english]{babel}
\usepackage{amsmath}
\usepackage{amssymb}
\usepackage{float}
\usepackage{graphicx}
\usepackage[colorinlistoftodos]{todonotes}
\usepackage{setspace}

\usepackage{hyperref}
\usepackage[margin=0.8in]{geometry}
\usepackage{pgfplots}
\pgfplotsset{compat=1.12}
\title{
 Variational Bayes Neural Network: Posterior Consistency, Classification Accuracy and Computational Challenges}
\author{Shrijita Bhattacharya\footnote{Corresponding author: {\rm bhatta61@msu.edu}}\: \footnote{Department of Statistics and Probability, Michigan State University} \quad Zihuan Liu $^\dagger$  \quad Tapabrata Maiti $^\dagger$}
\date{} 

\usepackage{tikz}
\usepackage{algorithm}
\usepackage[nottoc]{tocbibind}
\usepackage{wrapfig}

\newtheorem{theorem}{Theorem}[section]
\newtheorem{lemma}[theorem]{Lemma}
\newtheorem{corollary}[theorem]{Corollary}
\newtheorem{proposition}[theorem]{Proposition}
\newtheorem{definition}[theorem]{Definition}

\usepackage{booktabs}
\usepackage{multirow}

\newcommand{\pp}[2]{\frac{\partial{#1}}{\partial{#2}}}

\newcommand{\R}{\mathbb{R}}

\newcommand{\boldI}{\boldsymbol{I}}

\newcommand{\boldX}{\boldsymbol{X}}

\newcommand{\boldt}{\boldsymbol{t}}
\newcommand{\boldu}{\boldsymbol{u}}
\newcommand{\boldv}{\boldsymbol{v}}

\newcommand{\boldx}{\boldsymbol{x}}
\newcommand{\boldy}{\boldsymbol{y}}

\newcommand{\balpha}{\boldsymbol{\alpha}}
\newcommand{\bbeta}{\boldsymbol{\beta}}

\newcommand{\bgamma}{\boldsymbol{\gamma}}

\newcommand{\bomega}{\boldsymbol{\omega}}
\newcommand{\bzeta}{\boldsymbol{\zeta}}

\newcommand{\bTheta}{\boldsymbol{\Theta}}
\newcommand{\btheta}{\boldsymbol{\theta}}
\newcommand{\bmu}{\boldsymbol{\mu}}
\newcommand{\bpsi}{\boldsymbol{\psi}}
\newcommand{\bSigma}{\boldsymbol{\Sigma}}
\newcommand{\bxi}{\boldsymbol{\xi}}
\newcommand{\bGamma}{\boldsymbol{\Gamma}}

\begin{document}

\maketitle
\begin{abstract}
Bayesian neural network models (BNN) have re-surged in recent years due to the advancement of scalable computations and its utility in solving complex prediction problems in a wide variety of applications. Despite the popularity and usefulness of BNN, the conventional Markov Chain Monte Carlo based implementation suffers from high computational cost, limiting the use of this powerful technique in large scale studies. The variational Bayes inference has become a viable alternative to circumvent some of the computational issues. Although the approach is popular in machine learning, its application in statistics is somewhat limited.
This paper develops a variational Bayesian neural network estimation methodology and related statistical theory. The numerical algorithms and their implementational are discussed in detail. The theory for posterior consistency, a desirable property in nonparametric Bayesian statistics, is also developed. This theory provides an assessment of prediction accuracy and guidelines for characterizing the prior distributions and variational family. The loss of using a variational posterior over the true posterior has also been quantified.   The development is motivated by an important biomedical engineering application, namely building predictive tools for the transition from mild cognitive impairment to Alzheimer’s disease. The predictors are multi-modal and may involve complex interactive relations.

\end{abstract}
{\it Key Words: ADNI; Black-box variational inference, Kullback-Leibler divergence; Hellinger neighborhood;  Mean-field family.}

\section{Introduction}
Due to the universal approximation theory of stochastic functions and larger access to computational power, Bayesian Neural Networks (BNNs) are fashionable in machine learning and statistics for classification and prediction from big data. The BNNs based prediction have several advantages over standard parametric statistical models as they implicitly take into account the interactions or dependence among predictor variables and model the unknown functional relationship between the predictors and responses.  For example, we consider an application of classifying Alzheimer's disease status from brain imaging, an important biomedical problem. The image features are segmented into voxels or region of interests (ROI's). Due to their physical adjacency and other biological proximities, a simple parametric model or semi-parametric models, such as logistic regression or generalized additive models may not be appropriate. Besides the dependence (spatial) among the  predictors, there might be some network structures in the feature space while modeling the brain images. The BNNs can take into account these data features without any explicit assumptions about their dependence structure. Further, these studies often have additional features but in different modes such as genetic and demographic information, brings additional complexity while modeling dependence among the features. Thus, machine learning based approach, such as neural networks, becomes useful in this type of applications.
\par

Bayesian neural networks (BNNs) have been comprehensively studied by \cite{BIS1997}, \cite{Neal1992}, \cite{LAM2001}, and many others. More recent developments  which establish the efficacy of BNNs can be found in \cite{PML},  \cite{MULL2018}, \cite{ALI2018}, \cite{FML2018}, \cite{JAV2020} and the references therein. The estimation of posterior distribution is a key part of Bayesian inference and represents the information about the uncertainties for both data and parameters. However, exact analytical solution for the posterior distribution is intractable as the number of parameters is very large and the functional form of a neural network does not lend itself to exact integration \cite{blundell2015weight}. Several approaches have been proposed for solving posterior distribution of weights of BNNs, based on both  optimization based techniques such as variational inference (VI), and sampling based approach, such as Markov Chain Monte Carlo (MCMC). MCMC techniques are typically used to obtain sampling based estimates of the posterior distribution. Indeed, BNNs with MCMC have not seen widespread adoption due to computational cost in  terms  of  both  time and storage on a large dataset \cite{kingma2015variational,nagapetyan2017true,wan2018neural,wu2018deterministic}. In contrast to MCMC, VI tends to converge faster, and it has been applied to many popular Bayesian models, such as factorial models and  topic models \cite{li2018,Blei_2007,blei2003}.
We want to take a variational approximation approach for posterior estimation in the context of neural network predictive models.
The basic idea of VI is that it first defines a family of variational distributions, and then minimizes the Kullback-Leibler (KL) divergence with respect to the variational family. Many recent works have discussed the application of variational inference to Bayesian neural networks e.g.,  \cite{LG2009}, \cite{ALE2011}, \cite{CAR2012}, \cite{blundell2015weight}, \cite{SUN2019}. Although, there is a plethora of literature implementing  variational inference for neural networks, the theoretical properties of the variational posterior in BNNs remain relatively unexplored and this limits the use of this powerful computational tool beyond the machine learning community.  

Some of the previous works that focused on theoretical properties of variational posterior include  the frequentist  consistency of variational inference in parametric models in the presence of latent variables (see \cite{YD2019}).  Optimal risk bounds for mean-field variational Bayes for Gaussian mixture (GM) and Latent Dirichlet allocation (LDA) models have been discussed in \cite{PAT2017}. The work of \cite{YAN2017} propose $\alpha$-variational inference  Bayes risk for GM and LDA models.  A more recent work \cite{ZHA2017}  discusses the variational posterior consistency rates in Gaussian sequence models, infinite exponential families and piece-wise constant models.  In order to   evaluate the validity of a posterior in non-parametric models, one must establish its consistency and rates of contraction. To the best of our knowledge, for classification problems, the consistency rates of VI based Bayesian neural networks have not been explored. Additionally, there exists no previous work discussing the impact of using a VI approach on the classfication accuracy of the BNN's. Both these issues have been addressed in this paper. {\it As a systematic development of Variational Bayes Neural netowrks with theoretical rigor, our contributions are summarized as follows}: 

We first demonstrate that the variational posterior asymptotically gives negligible probability to shrinking Hellinger neighborhoods of the true density function provided that the rate of growth in the number of nodes is well controlled.


Besides the theoretical validation, the challenges of implementing a VI based approach is two folds: (1) the optimization of the KL-divergence (2) the choice of the variational family. There exists several algorithms for the optimization of the KL-divergence measure, each having its own pros and cons. We implemented the black-box variational inference (BBVI) to obtain Monte Carlo estimates of the gradient of the evidence lower bound (ELBO) and used stochastic optimization to estimate the variational parameters. We reveal that the complexity involved in this optimization depends on several factors and tuning parameter selection. We then provide statistically principled guideline in couple with optimization theory for its practical implementation. For the second issue, we show that a simple mean-field variational family suffices for posterior consistency along with good numerical performance. 

Further we derive the rates of consistency for the true posterior and the variational posterior. With no assumptions on the true function, we first establish that the true posterior probability of an $\varepsilon$- Hellinger neighborhood  grows at the rate $1-2e^{-n\varepsilon^2/2}$ in contrast to the slower $1-\nu, 0<\nu<1$ rate for the variational posterior. Next, under suitable assumptions on the approximating neural network solution of the true function, we establish that the true posterior probability of a shrinking $\varepsilon \epsilon_n$- Hellinger neighborhood  grows at the rate $1-2e^{-n\varepsilon^2 \epsilon_n^2/2}$ in contrast to the slower $1-\nu$ rate for the variational posterior. The reason for this difference is explained as follows: (1) first, the  true posterior probability of a shrinking $\varepsilon \epsilon_n$- Hellinger neighborhood  grows at the rate $1-2e^{-n\varepsilon^2\epsilon_n^2/2}$ (2) second, the KL-distance between the variational posterior and the true posterior grows at a rate  $o(n\epsilon_n^2)$ . For (1) and (2) to simultaneously hold, for every $\nu>0$, the variational posterior must give greater than $1-\nu$ probability to shrinking Hellinger neighborhoods otherwise the rate of growth of the KL-distance between the true and variational posterior cannot be controlled. Indeed, for (1) to hold, one must choose the prior appropriately and for (2) to hold, one must choose the variational family appropriately. Indeed, for the optimal classification, one needs to optimally choose the  pair of prior and variational family. This paper carefully characterizes the properties of a prior and variational family which allows the variational posterior to be consistent and establishes that a simple mean field family choice for both these families provides a numerically stable and a theoretically consistent solution.

We next establish the connection between posterior consistency and classification accuracy. In this direction, we first show that the classification accuracy of a consistent posterior asymptotically approaches to the classification accuracy of a Bayes classifier. With no assumption on the true function, we show that the rates of convergence of the classification accuracy is same for both variational approximation and true posterior. However, under suitable assumptions on the approximating neural network solution, we establish that the classification accuracy of variational posterior approaches to the classification accuracy of the Bayes classification at the rate $(\epsilon_n^2)^{2/3}$ in contrast to the higher rate of $\epsilon_n^2$ for the true posterior. This interesting theoretical discovery quantifies the loss due to to the use of variational posterior instead of using the true posterior density. 

Finally, we discuss the usefulness of this modeling technique for prediction from a multimodal data that characterizing various aspects among the feature variables. Specifically, we provide an application in the context of Alzheimer disease classification from brain imaging, genetic variable and other clinical factors.

The rest of the paper is organized as follows:  section \ref{Section:Model} presents the statistical framework for neural networks based classification problem and section \ref{sec:bayes} provides the variational algorithm  for Bayesian implementation.
The theoretical properties including posterior consistency and classification consistency are provided in section \ref{Sec:Theory}.
The detailed numerical issues in practice are explained in section \ref{Sec:Numerical} in the context of a real life biomedical engineering application. Section \ref{Sec:Discussion} concludes the article with brief remarks on research directions.

\section{The Neural Networks Classifier and Likelihoods}
\label{Section:Model}

 Let $Y$ be a binary random variable taking values 0 or 1, representing the class levels and $X\in\R^p$ is a feature vector drawn from a feature space with some marginal distribution $P_{X}$. We consider the following binary classification problem \begin{equation}
\label{e:class}
    P(Y=1|X=x)=\sigma(\eta_0(\boldx)),\ P(Y=0|X=x)=1-\sigma(\eta_0(\boldx))
\end{equation}
where $\eta_0(\cdot):\R^p\rightarrow\R$ is some continuous function and $\sigma(.)=e^{(.)}/(1+e^{(.)})$ is the sigmoid function. Thus, $P_{X,Y}$, the joint distribution of $(X,Y)$ is a product of the conditional distribution in \eqref{e:class} and the marginal distribution $P_{X}$. Borrowing some notations from \cite{CANN} and \cite{yangmaiti2020}, a classifier $C$ is a Borel
measurable function $C :\mathbb{R}^p \to\{0, 1\}$, with the interpretation that we assign a point $\boldx\in \mathbb{R}^p$ to class $C(\boldx)$. The test error of a classifier $C$ is given by
\begin{equation}
\label{e:riskintegral}
R(C)=\int_{\mathbb{R}^p\times\{0,1\}}I_{\{C(X)\neq Y\}}dP_{X,Y}
\end{equation}
Based on \eqref{e:class}, we define the Bayes classifier as
\begin{equation}
\label{e:Bayes}
    C^{\rm Bayes}(\boldx)=\begin{cases} 
1, & \sigma(\eta_0(\boldx))\geq 1/2\\
0, & \text{otherwise}
\end{cases}
\end{equation}
The Bayes classifier is optimal \cite{friedman2001elements} since it minimizes the mis-classification error risk in  \eqref{e:riskintegral}. However, the Bayes classifier is not useful in practice, since the function $\eta_0(\boldx)$ is unknown. Thus, a classifier is  obtained based on a set of training observations $\{(\boldx_1,y_1),...,(\boldx_n,y_n)\}$, which are drawn from $P_{X,Y}$. A good classifier based on the sample should have the risk tending to the Bayes risk as the number of observations tends to infinity, without any requirement for its probability distribution. This is so called universal consistency.
Multiple methods have been adopted to estimate $\eta_0(\boldx)$, including logistic regression (a linear approximation), generalized additive model (GAM, a nonparametric nonlinear approximation), neural networks (a complicated structure which is dense in continuous functions)  etc.

\begin{figure}[h]
\begin{tikzpicture}[
plain/.style={
  draw=none,
  fill=none,
  },
net/.style={
  matrix of nodes,
  nodes={
    draw,
    circle,
    inner sep=6pt
    },
  nodes in empty cells,
  column sep=2cm,
  row sep=-8pt
  },
>=latex
]
\matrix[net] (mat)
{
|[plain]| \parbox{1.8cm}{\centering Input\\layer} & |[plain]| \parbox{1.8cm}{\centering Hidden\\layer} & |[plain]| \parbox{1.8cm}{\centering Output\\layer} \\
& |[plain]| \\
|[plain]| & \\
& |[plain]| \\
  |[plain]| & |[plain]| \\
|[plain]|... & |[plain]|...& \\
  |[plain]| & |[plain]| \\
& |[plain]| \\
  |[plain]| & \\
& |[plain]| \\    };
\foreach \ai [count=\mi ]in {2,4}
  \draw[<-] (mat-\ai-1) -- node[above] {Input \mi} +(-3.5cm,0);
\foreach \ai [count=\mi ]in {8}
  \draw[<-] (mat-\ai-1) -- node[above] {Input $p-1$} +(-3.5cm,0);
\foreach \ai [count=\mi ]in {10}
  \draw[<-] (mat-\ai-1) -- node[above] {Input $p$} +(-3.5cm,0);
\foreach \ai in {2,4,8,10}
{\foreach \aii in {3,6,9}
  \draw[->] (mat-\ai-1) -- (mat-\aii-2);
}
\foreach \ai in {3,6,9}
  \draw[->] (mat-\ai-2) -- (mat-6-3);
\draw[->] (mat-6-3) -- node[above] {Ouput} +(2cm,0);
\end{tikzpicture}
\caption{A diagram for the single-layer neural network model}
\label{fullnn}
\end{figure}
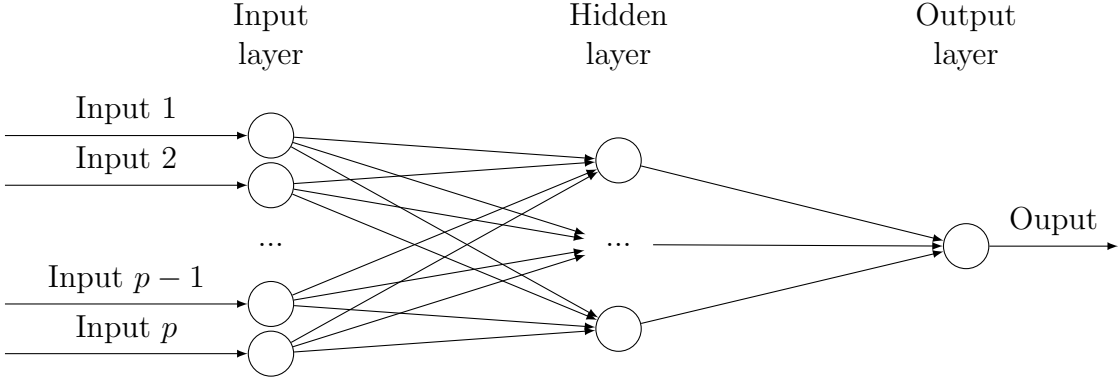

The first two methods usually work in practice with good theoretical foundation, however, they may fail to catch the complicated dependency of the feature vector $\boldx$ in a wide range of applications including the problem that we consider in this article. The neural network structure exploits the dependency implicitly without any specific parametric structure. Consider a single layer neural network model with $p$ predictor variables. The layer has $k_n$ nodes, where $k_n$ may be a diverging sequence depending on $n$. 
A diagram is shown in figure \ref{fullnn} for illustration.

The validity of neural network approximations is based on  the universal approximation results \cite{cybenko1989approximation}, which states that  the single layer neural network is able to approximate any continuous function with a quite small approximation error when $k_n$ is large.  Assume a Fourier representation of $\eta_0(\boldx)$ 
and denote $\Gamma_{B,C}=\{f(\cdot):\int_{B}\|\bomega\|_2|\tilde{f}|(d\bomega)<C\}$ for some bounded subset $B$ of $\R^p$ containing zero for some constant $C>0$. Then, for all functions $\eta_0\in\Gamma_{B,C}$, there exist a single layer neural network output $\eta(\boldx)$ such that $\|\eta-\eta_0\|_2=O(1/\sqrt{k_n})$  \cite{barron1993universal}. This result ensures  good approximation property of single layer neural network, and the convergence rate depends only on the number of nodes under mild conditions on $\eta_0(\boldx)$.

For the $p\times 1$ input vector $\boldx$, and weight $\bgamma$,  let $k_n\times 1$ vector $\bxi$ be the corresponding values in the hidden nodes, i.e.,
$$\xi_j=\gamma_{j0}+\bgamma_{j}^{\top}\boldx,\ j=1,...,k_n$$
Let $\psi(\cdot)$ be an activation function, then the output for a given set of weight $\bbeta$, is calculated by
$$\beta_0+\bbeta^T\bpsi(\bxi)$$
where the function $\bpsi(\cdot)$ is the function $\psi(\cdot)$ being applied element-wise.  \cite{leshno1993multilayer} proved that as long as the activation function is not algebraic polynomials, the single layer neural network is dense in the continuous function space, thus can be used to approximate any given continuous function. This structure can be considered as a non-parametric model.  For a given activation function $\psi(\cdot)$,  a $p\times 1$ input vector maps to an output score
\begin{eqnarray}
\label{e:eta-x}
\nonumber    \eta_{\btheta_n}(\boldx)&=&\beta_0+\sum_{j=1}^{k_n}\beta_j\psi(\gamma_{j0}+\bgamma^T_j\boldx)\\
&=&\beta_0+\bbeta^\top \psi(\bgamma_0+\bGamma\boldx)
\end{eqnarray}
where $\bgamma_0=[\gamma_{10}, \cdots, \gamma_{k_n0}]$ and $\bGamma=[\bgamma_1, \cdots, \bgamma_{k_n}]$ and $\btheta_n=(\beta_0, \bbeta, \bgamma_0, {\rm vec}(\bGamma))$ is the set of all the parameters. Note, $\btheta_n$ is a $K(n)\times 1$ vector where $K(n)=k_n(p+1)+1$. Assuming the number of covariates $p$ to be fixed, the total number of parameters $K(n) \sim k_n$, where $k_n$ is the number of nodes. For the purposes of this paper, we use the activation function to be the sigmoid function, $$\psi(x)=\frac{e^x}{1+e^x}.$$
although the results may be generalized to a wider class of activation functions. Thus, using the neural network in \eqref{e:eta-x} as an approximation to the true function $\eta_0(\boldx)$ in \eqref{e:class},  the conditional probabilities of $Y$ given $X=\boldx$ is given by
\begin{eqnarray}
\label{e:sig-x}
\nonumber    P(Y=1|X=\boldx)&=& \sigma(\eta_{\btheta_n}(\boldx))\\ 
P(Y=0|X=\boldx)&=& 1-\sigma(\eta_{\btheta_n}(\boldx))
\end{eqnarray}
Assuming Bernoulli probability distribution, the conditional density function of $Y|X=\boldx$ under the model is given by
\begin{equation}
\label{e:lik-one}
\ell_{\btheta_n}(y,\boldx)=\exp\left(y \eta_{\btheta_n}(\boldx)-\log\left(1+e^{ \eta_{\btheta_n}(\boldx)}\right)\right)
\end{equation}
Thus, the likelihood function for the data $(\boldy_n,\boldX_n)=(y_i,\boldx_i)_{i=1}^n$ under the model can be expressed as.
\begin{equation}
\label{e:lik}
L(\btheta_n)=\prod_{i=1}^n \ell_{\btheta_n}(y_i,\boldx_i)=\exp\left(\sum_{i=1}^n\left[y_i \eta_{\btheta_n}(\boldx_i)-\log\left(1+e^{ \eta_{\btheta_n}(\boldx_i)}\right)\right]\right)
\end{equation}
In view of \eqref{e:class}, the conditional density of  $Y|X=\boldx$ under the truth
 \begin{equation}
 \label{e:lik-one-0}
 \ell_{0}(y,\boldx)=\exp\left(y \eta_{0}(\boldx)-\log\left(1+e^{ \eta_{0}(\boldx)}\right)\right)
 \end{equation}
Therefore, the likelihood function for the data under the truth is given by
\begin{equation}
\label{e:lik-0}
 L_0=\prod_{i=1}^n \ell_0(y_i,\boldx_i)=\exp\left(\sum_{i=1}^n\left[y_i \eta_0(\boldx_i)-\log\left(1+e^{ \eta_0(\boldx_i)}\right)\right]\right)
\end{equation}

\section{Bayesian Inference with Variational Algorithm}
\label{sec:bayes}

\subsection{Prior Choice}
For Bayesian analysis, prior distributions have to be assigned for all parameters defining the model. Although one may have a prior knowledge concerning the function represented by a neural network, it is generally difficult to translate this into a meaningful prior on neural network weights. 

We assume a normal prior on each entry of $\bTheta_n$ as follows:
\begin{equation}
\label{e:prior}
 p(\btheta_n)=\prod_{j=1}^{K(n)}\frac{1}{\sqrt{2\pi \zeta_{jn}^2}}e^{-\frac{1}{2\zeta_{jn}^2} (\theta_{jn}-\mu_{jn})^2}
\end{equation}
where each $\Theta_{jn}$ requires a separate mean $\mu_{jn}$ and variance $\zeta^2_{jn}$. 
Note, we use $\bTheta_n$ to denote the random variable for parameter $\btheta_n$.

Let $\bzeta_n=[\zeta_{1n}, \cdots, \zeta_{K(n) n}]$ and $\bzeta_n^*=[1/\zeta_{1n}, \cdots, 1/\zeta_{K(n) n}]$, we assume the following two conditions on the variance parameter
\begin{eqnarray*}
\hspace{-40mm}{\bf (A1)} \hspace{40mm}  ||\bzeta_n||_\infty&=&O(n)\\
||\bzeta^*_n||_\infty&=&O(1)
\end{eqnarray*}
where $||.||_\infty$ is the supremum norm of a vector as in definition \ref{def:norms}.  Note, the above assumption ensures that the variances of each $\bTheta_n$ do not grow at an arbitrarily large rate in which case the consistency of both the Bayesian and variational Bayes approach would break down. Restrictions on the mean parameter $\bmu_n=[\mu_{1n},\cdots, \mu_{K(n)n}]$ directly impact the consistency rate and are more case specific. We thereby leave their thorough discussion under section \ref{Sec:Theory}. 

The reason for choosing the above form of prior is two folds: (1) first it guarantees that the true posterior distribution of $\bTheta_n$  is consistent (2) second it guarantees, under a suitable choice of the variational family, the approximated variational posterior is also consistent. The choice of prior in \eqref{e:prior} is not unique. Indeed, one can work with a much more generic class of prior such that (1) and (2) hold. Note, each prior comes with its own associated computation complexity, implementation and theoretical justification. We choose one which does a fairly good job under all these three criterion.

In view of \eqref{e:lik} and \eqref{e:prior}, posterior distribution of $\bTheta_n$ given $\boldy_n=[y_1,\cdots, y_n]^\top$ and $\boldX_n=[\boldx_1,\cdots, \boldx_n]^\top$ is
\begin{equation}
\label{e:posterior}
\pi(\btheta_n|\boldy_n,\boldX_n)=\frac{\pi(\btheta_n,\boldy_n,\boldX_n)}{\pi(\boldy_n,\boldX_n)}=\frac{L(\btheta_n)p(\btheta_n)}{\int L(\btheta_n)p(\btheta_n)d\btheta_n}
\end{equation}
where the denominator $\pi(\boldy_n,\boldX_n)$ is free from the parameter and depends only on $\boldy_n$ and $\boldX_n$.

\subsection{Variational Inference}


As a first step to variational inference  (VI) procedure, one has to start with a variational family. Given several options, we work with one which is simple, computationally and structurally tractable, and more importantly statistically consistent. We posit a mean field Gaussian variational family of the form
\begin{equation}
\label{e:VIF}
\mathcal{Q}_n=\left\{q(\btheta_{n}): q(\btheta_{n})=\prod_{j=1}^{K(n)}\frac{1}{\sqrt{2\pi s^2_{jn}}}e^{-\frac{1}{2s_{jn}^2}(\theta_{jn}-m_{jn})^2}\right\}
\end{equation}
Note that the variational family assumes that each $\Theta_{jn}$ is independent with mean and standard deviation equal to $m_{jn}$ and $s_{jn}$ respectively. 

The variational posterior aims to reduce the KL-distance between the variational family and the true posterior \cite{Blei_2007,Hinton93,zhang2017noisy,blundell2015weight}.
Thus, for the true posterior, $\pi(.|\boldy_n,\boldX_n)$ in \eqref{e:posterior}, the variational posterior is given by
\begin{equation}
\label{e:var-posterior}
    \pi^*=\underset{q \in \mathcal{Q}_n}{\text{argmin}}\:\: d_{\rm KL}(q, \pi(.|\boldy_n,\boldX_n)).
\end{equation}
 where $d_{\rm KL}$, the Kullback-Leibler (KL) divergence  between a variational family member $q(\boldsymbol{\theta_n})$ and the true posterior $\pi(\btheta_n|\boldy_n,\boldX_n)$ is given by 
\begin{equation}
\label{e:KL_qp}
d_{\rm KL}(q, \pi(.|\boldy_n,\boldX_n))=\int \log \frac{ q(\btheta_n)}{\pi(\btheta_n|\boldy_n,\boldX_n)} q(\btheta_n) d\btheta_n
\end{equation}
From \eqref{e:posterior}, note that $\pi(\btheta_n|\boldy_n,\boldX_n)=\pi(\btheta_n,\boldy_n,\boldX_n)/\pi(\boldy_n,\boldX_n)$. Thus, simplifying the KL-divergence further, we get:
\begin{align}
\label{e:KL_qp_1}
\nonumber d_{\rm KL}(q, \pi(.|\boldy_n,\boldX_n))&=\int [\log q(\btheta_n)-\log \pi(\btheta_n,\boldy_n,\boldX_n)+\pi(\boldy_n,\boldX_n))] q(\btheta_n) d\btheta_n\\
&=\int [\log q(\btheta_n)-\log \pi(\btheta_n,\boldy_n,\boldX_n)] q(\btheta_n) d\btheta_n+\log \pi(\boldy_n,\boldX_n)
\end{align}
Since the last term in \eqref{e:KL_qp_1} does not depend $q$, optimizing \eqref{e:KL_qp_1} w.r.t. to $q$ boils down to optimizing the first term. Indeed the first term is nothing but the negative of the evidence lower bound (ELBO) where the ELBO is given by
\begin{equation}
    \label{e:elbo}
    \text{ELBO}(q,\pi(.,\boldy_n,\boldX_n))=\int  [\log \pi(\btheta_n,\boldy_n,\boldX_n)-\log q(\btheta_n)] q(\btheta_n) d\btheta_n 
\end{equation}
Thus in order to minimize the KL-distance, we shall instead maximize the ELBO between $q$ and $\pi(.,\boldy_n,\boldX_n)$.  For the purposes of implementation, we alternatively define $\pi^*$ as
\begin{equation}
\label{e:var-posterior-n}
    \pi^*=\underset{q \in \mathcal{Q}_n}{\text{argmax}}\:\: \text{ELBO}(q,\pi(.,\boldy_n,\boldX_n))
\end{equation}
In order to maximize ELBO in \eqref{e:elbo} with respect to the distribution $q$ we index each density $q$ in the variational family $\mathcal{Q}_n$ by its paramters,
\begin{equation}
\label{e:var-par}
\begin{split}
\mathcal{V}_q=(m_{1n},\cdots,m_{K(n) n},s^2_{1 n},\cdots,s^2_{K(n) n})
\end{split}
\end{equation}
where $m_{jn}$ and $s_{jn}$ is the mean and standard deviation of $\Theta_{jn}$ under the density $q$. This consequently produces an indexing of the \text{ELBO} in \eqref{e:elbo} as
\begin{equation}
    \label{e:elbo-idx}
\mathcal{L}_{\mathcal{V}_q}  =  \text{ELBO}(q(.|\mathcal{V}_q),\pi(.,\boldy_n,\boldX_n))=\int  [\log \pi(\btheta_n,\boldy_n,\boldX_n)-\log q(\btheta_n|\mathcal{V}_q)] q(\btheta_n|\mathcal{V}_q) d\btheta_n 
\end{equation}
Thus, the optimization problem in \eqref{e:var-posterior-n} reduces to optimizing $\mathcal{L}_{\mathcal{V}_q}$ w.r.t. to $\mathcal{V}_q$. In this direction, we consider the derivative of $\mathcal{L}_{\mathcal{V}_q}$ w.r.t. $\mathcal{V}_q$. Thus,
\begin{align}
\label{e:grad-simp}
 \nonumber   \nabla_{\mathcal{V}_q} \mathcal{L}_{\mathcal{V}_q}&=\nabla_{\mathcal{V}_q} \int  [\log \pi(\btheta_n,\boldy_n,\boldX_n)-\log q(\btheta_n|\mathcal{V}_q)] q(\btheta_n|\mathcal{V}_q) d\btheta_n \\
\nonumber    &=\int   \nabla_{\mathcal{V}_q}  q(\btheta_n|\mathcal{V}_q) d\btheta_n+\int    [\log \pi(\btheta_n,\boldy_n,\boldX_n)-\log q(\btheta_n|\mathcal{V}_q)] \nabla_{\mathcal{V}_q} q(\btheta_n|\mathcal{V}_q) d\btheta_n\\
    &=\int   \nabla_{\mathcal{V}_q}  \log q(\btheta_n|\mathcal{V}_q)  [\log \pi(\btheta_n,\boldy_n,\boldX_n)-\log q(\btheta_n|\mathcal{V}_q)]  q(\btheta_n|\mathcal{V}_q)  d\btheta_n
\end{align}
where the last equality follows since $\int   \nabla_{\mathcal{V}_q}  q(\btheta_n|\mathcal{V}_q) d\btheta_n = \nabla_{\mathcal{V}_q}  \int   q(\btheta_n|\mathcal{V}_q)d\btheta_n=\nabla_{\mathcal{V}_q} 1=0$ and $\nabla_{\mathcal{V}_q}  \log q(\btheta_n|\mathcal{V}_q) q(\btheta_n|\mathcal{V}_q)=\nabla_{\mathcal{V}_q} q(\btheta_n|\mathcal{V}_q)$. Therefore, the gradient of $\mathcal{L}_{\mathcal{V}_q}$ w.r.t. $\mathcal{V}_q$ reduces to 
\begin{equation}
    \label{e:grad}
     \nabla_{\mathcal{V}_q} \mathcal{L}_{\mathcal{V}_q}=E_{q(.|\mathcal{V}_q)}[\nabla_{\mathcal{V}_q}  \log q(\bTheta_n|\mathcal{V}_q)  [\log \pi(\bTheta_n,\boldy_n,\boldX_n)-\log q(\bTheta_n|\mathcal{V}_q)] ]
\end{equation}

The black-box variational inference (BBVI) algorithm  \cite{ranganath2013black}, optimizes the ELBO using gradient descent method by making use the above form of  gradient. Note,  the key challenge in evaluating the gradient in \eqref{e:grad} is the computation of the expectation. Exact computation of the expectation leads to high computational complexity whereas using noisy estimates leads to high variability. In section \ref{sec:BBVI}, we elucidate how to balance the trade off between these two factors in order to ensure fast and efficient estimation of the gradient.





\subsection{Black Box Variational Algorithm using score function estimator}

\label{sec:BBVI}

The black box variational inference (BBVI) algorithm aims at computing the derivative in \eqref{e:grad} in an efficient manner. 
The gradient (\ref{e:grad}) is difficult to evaluate for problems with complex likelihood structures arising out of neural network models. Alternatively, the above expectation is evaluated using Monte Carlo techniques—by sampling from the variational distribution and forming the corresponding Monte Carlo estimates of the gradient.
These Monte Carlo estimates can be then used for  stochastic optimization to fit the variational parameters (see \cite{ranganath2013black,Jalil}). The Monte Carlo estimator of the quantity in (\ref{e:grad}), obtained by sampling from the variational distribution is referred to as the score-function estimator. Indeed, the score function estimator technique views the gradient of ELBO as the expectation with respect to the variational distribution using the log-derivative trick (see the derivation in \eqref{e:grad-simp}). This technique implicitly assumes that the Lebesgue’s dominated convergence theorem is applicable so that one can take the gradient of the expectation in by moving the gradient inside the expectation (see \cite{ranganath2013black,mcbook,Jalil}). 

We next explain the computation of Monte Carlo estimate of the gradient in  (\ref{e:grad}). In this direction, let $\mathcal{V}_q$ denote the current value of the variational parameters. We generate $S$ samples from the variational distribution $q(.|\mathcal{V}_q)$. We then define the noisy but unbiased estimate of the gradient as 
\begin{equation}
    \widehat{\nabla \mathcal{L}}_{\mathcal{V}_q}=\frac{1}{S} \sum_{w=1}^S \nabla_{\mathcal{V}_q} \log q (\btheta_n[w]|\mathcal{V}_q)[\log \pi(\btheta_n[w],\boldX_n,\boldy_n)
     -\log q (\btheta_n[w]|\mathcal{V}_q)]
     \label{e:mcmcelbo-0}
\end{equation}
where $\btheta_n[1],\cdots, \btheta_n[S]$ are samples generated from $q (.|\mathcal{V}_q)$. Analogous to \eqref{e:mcmcelbo-0}, a noisy but unbiased estimate of the ELBO in \eqref{e:elbo} is given by    \begin{equation}
   \widehat{\mathcal{L}}_{\mathcal{V}_q}=\frac{1}{S} \sum_{w=1}^S[\log \pi(\btheta_n[w],\boldX_n,\boldy_n)
     -\log q (\btheta_n[w]|\mathcal{V}_q)]
   \label{e:elbo-val}
    \end{equation}

Algorithm~\ref{al:bbvi} provides the pseudocode summarizing the overall algorithm for training BBVI using score function estimator.

\begin{spacing}{0.3}
\begin{algorithm}[H]
\begin{enumerate}
    \item Fix an initial value for variational family parameters $\mathcal{V}_q^1$.
    \item Fix a step size sequence $\rho_t$, $t=1, \cdots$.
    \item Set $t=1$.
    \item Simulate $S$ samples $\btheta_n[1], \cdots, \btheta_n[S]$  from $q(.|\mathcal{V}_q^t)$.
    \item Compute  \begin{equation*}
    \widehat{\nabla \mathcal{L}}_{\mathcal{V}_q^t}=\frac{1}{S} \sum_{w=1}^S \nabla_{\mathcal{V}_q^t} \log q (\btheta_n[w]|\mathcal{V}_q^t)[\log \pi(\btheta_n[w],\boldX_n,\boldy_n)
     -\log q (\btheta_n[w]|\mathcal{V}_q^t)]
     \label{e:mcmcelbo}
\end{equation*}
    \item Update \begin{equation}
        \mathcal{V}_q^{t+1}=\mathcal{V}_q^{t}+\rho_t  \widehat{\nabla \mathcal{L}}_{\mathcal{V}_q^t}
        \label{e:grad-update}
    \end{equation}
    \item Set $t=t+1$.
    \item Repeat steps 4-7 until the convergence of  $\widehat{\mathcal{L}}_{\mathcal{V}_q^t}$ where
    \begin{equation*}
   \widehat{\mathcal{L}}_{\mathcal{V}_q^t}=\frac{1}{S} \sum_{w=1}^S[\log \pi(\btheta_n[w],\boldX_n,\boldy_n)
     -\log q (\btheta_n[w]|\mathcal{V}_q^t)]
   \label{e:elbo-val-1}
    \end{equation*}
\end{enumerate}
\caption{BBVI using score function estimator}
\label{al:bbvi}
\end{algorithm}
\end{spacing}

Note the most crucial step in the implementation of algorithm \ref{al:bbvi} is the computation of the quantity $\nabla_{\mathcal{V}_q} \log q(\btheta_n|\mathcal{V}_q)$ for the variational parameters $\mathcal{V}_q$ as in \eqref{e:var-par}. For the choice of $q$ as in \eqref{e:VIF} and the variational parameters $(m_{1n},\cdots, m_{K(n)n})$, the explicit expressions for $\nabla_{\mathcal{V}_q} \log q(\btheta_n|\mathcal{V}_q)$  have been presented in appendix (A) (see section \ref{sec:append}). 

For the variational parameters $(s_{1n}, s_{2n}, \cdots, s_{K(n)n})$, the updating rule in \eqref{e:grad-update} may lead to negative estimates of $s_{jn}$, $j=1, \cdots, K(n)$. However, one must guard against this since variance terms cannot be negative. Thus, to perform the optimization,  we reparametrize the variance terms as $s_{jn}=\log (1+e^{r_{jn}})$, $j=1, \cdots, K(n)$ and update the quantities $r_{jn}$ in each step instead of $s_{jn}$. Note, by chain rule, 
\begin{equation}
\label{e:chain-rule}
    \nabla{r_{jn}}\log q(\btheta_n|\mathcal{V}_q)=\left(\frac{e^{r_{jn}}}{1+e^{r_{jn}}}\right)\left( \nabla{s_{jn}}\log q(\btheta_n|\mathcal{V}_q)\Big|_{s_{jn}=\log (1+e^{r_{jn}})}\right) 
\end{equation}
where the term in the second bracket is the derivative of $\log q(\btheta_n|\mathcal{V}_q)$ with respect to (w.r.t.) $s_{jn}$ evaluated at the point $s_{jn}=\log(1+e^{r_{jn}})$ and the first term is the derivative of $s_{jn}$ w.r.t. $r_{jn}$. The explicit expressions for derivative of $\log q(\btheta_n|\mathcal{V}_q)$ w.r.t. to both $r_{jn}$ and $s_{jn}$ have been presented in the appendix (A) (see section \ref{sec:append}).


\subsection{Control Variate: Stabilizing the stochastic gradient}

\label{sec:cv}
We can use algorithm \ref{al:bbvi} to maximize the ELBO, however a major drawback is that the noisy estimator of the gradient has high variance. 
     There are two major techniques to reduce the variance of gradients. One of them is ``Rao-Blackwellization", where the idea  is to replace the noisy estimate of gradient with its conditional expectation with respect to a subset of the variables, \cite{ranganath2013black}. This method is useful when the posterior distribution is separable across subsets of variables especially while dealing with latent variables. However, a convoluted likelihood as in \eqref{e:lik} is not separable across the components of $\bTheta_n$ and there are no latent variables in our model. We thereby refrain from using the Rao-Blackwellization approach of gradient stabilization. 

The other method which also gives an efficient technique for stabilizing the gradient is called control variate (CV). Recently, control variates have been of interest for variational inference  and for general optimization problems that occur in machine learning \cite{ross2002,paisley2012variational,ranganath2013black}.
CV can be used to further reduce the variance of the MC approximations of the gradients. The key idea behind the variance reduction as proposed in \cite{ross2002} is to replace the target function, whose expectation is being approximated by Monte Carlo, with an auxiliary function that has the same expectation but a smaller variance. As described in both \cite{ross2002} and \cite{ranganath2013black}, to reduce the variance of the function $\xi(\phi)$, one instead considers the function $\hat \xi(\phi)=\xi(\phi)-a\left(\varphi(\phi)-E_q(\varphi(\phi))\right)$ where $\varphi(\phi)$ is function with finite expectation and $a$ is a scalar. Such a choice ensures $E_q(\hat \xi(\phi)) = E_q(\xi(\phi))$ and
$ \text{Var}_q(\hat \xi(\phi))= \text{Var}_q(\xi(\phi))+a^2\text{Var}_q(\varphi(\phi))-2a\text{Cov}_q(\xi(\phi),\varphi(\phi))$ which  is minimized at
\begin{equation}
a^\star=\frac{\text{Cov}_q(\xi(\phi),\varphi(\phi))}{\text{Var}_q(\varphi(\phi))}
\label{e:cv:a}
\end{equation}
Thus, greater the correlation between $\xi$ and $\varphi$, the greater the variance reduction. In the context of BBVI, \cite{ranganath2013black} proposed the use of $\nabla_{\mathcal{V}_q}\log q(\btheta|\mathcal{V}_q)$ as a choice for $\varphi(\phi)$. The stochastic approximation of the gradient in \eqref{e:mcmcelbo-0} can then modified as
\begin{equation}
  \widehat{\nabla \mathcal{L}}_{\mathcal{V}(H)}=\frac{1}{S} \sum_{w=1}^S \nabla_{\mathcal{V}_q} \log q (\btheta_n[w]|\mathcal{V}_q)[\log \pi (\btheta_n[w],\boldX_n,\boldy_n)
     -\log q (\boldsymbol{\theta_n[w]}|\mathcal{V}_q)-a^\star]
     \label{e:cv_optimazation}
\end{equation}
Since it is impossible to obtain an exact expression for $a^\star$ as in \eqref{e:cv:a}, one thus uses
\begin{equation}
    \label{e:a-hat-star}
    \widehat{a^\star}=\frac{\text{cov}(\boldu,\boldv)}{\text{var}(\boldv)}
\end{equation}
where both $\boldu$ and $\boldv$ are $S\times 1$ vectors whose $w^{\text{th}}$ element is given by
\begin{align}
    \label{e:u-v}
\nonumber    u[w]&= \nabla_{\mathcal{V}_q} \log q (\btheta_n[w]|\mathcal{V}_q)[\log \pi (\btheta_n[w],\boldX_n,\boldy_n)
     -\log q (\btheta_n[w]|\mathcal{V}_q)]\\
     v[w]&= \nabla_{\mathcal{V}_q} \log q (\btheta_n[w]|\mathcal{V}_q)
\end{align}

The extension of algorithm \ref{al:bbvi} with variance reduction of MC approximations due to CV is, is annotated as BBVI-CV and  summarized in algorithm \ref{al:bbvi-cv}.

Similar to the implementation of algorithm \ref{al:bbvi}, for the implementation of algorithm \ref{al:bbvi-cv}, we use the reparametrization of $s_{jn}=\log (1+e^{r_{jn}})$ as explained in section \ref{sec:BBVI}. We have implemented both the algorithms in our numerical study and discussed further practical complexity.

\begin{spacing}{0.05}
\begin{algorithm}[H]
\begin{enumerate}
        \item Fix an initial value for variational parameter $\mathcal{V}_q^1$.
    \item Fix a step size sequence $\rho_t$, $t=1, \cdots$.
    \item Set $t=1$.
    \item Simulate $S$ samples $\btheta_n[1], \cdots, \btheta_n[S]$  from $q(.|\mathcal{V}_q^t)$.
    \item Compute
    \begin{align*}
    \label{e:u-v}
\nonumber    u^t[w]&= \nabla_{\mathcal{V}_q^t} \log q (\btheta_n[w]|\mathcal{V}_q^t)[\log \pi (\btheta_n[w],\boldX_n,\boldy_n)
     -\log q (\btheta_n[w]|\mathcal{V}_q^t)]\\
     v^t[w]&= \nabla_{\mathcal{V}_q^t} \log q (\btheta_n[w]|\mathcal{V}_q^t)
\end{align*}
and 
    \begin{equation*}
    \label{e:a-hat-star}
    \widehat{a^\star}^t=\frac{\text{Cov}(\boldu^t,\boldv^t)}{\text{Var}(\boldv^t)}
\end{equation*}
    \item Compute  \begin{equation*}
    \widehat{\nabla \mathcal{L}}_{\mathcal{V}_q^t}=\frac{1}{S} \sum_{s=1}^S u^t[w]
     -\widehat{a^\star}^t v^t[w]
     \label{e:mcmcelbo}
\end{equation*}
    \item Update \begin{equation}
        \mathcal{V}_q^{t+1}=\mathcal{V}_q^{t}+\rho_t  \widehat{\nabla \mathcal{L}}_{\mathcal{V}_q^t}
    \end{equation}
    \item Set $t=t+1$.
    \item Repeat steps 4-7 until the convergence of  $\widehat{\mathcal{L}}_{\mathcal{V}_q^t}$ where
    \begin{equation*}
   \widehat{\mathcal{L}}_{\mathcal{V}_q^t}=\frac{1}{S} \sum_{w=1}^S[\log \pi(\btheta_n[w],\boldX_n,\boldy_n)
     -\log q (\btheta_n[w]|\mathcal{V}_q^t)]
   \label{e:elbo-val-2}
    \end{equation*}
\end{enumerate}
\caption{BBVI with control variates using score function estimator}
\label{al:bbvi-cv}
\end{algorithm}
\end{spacing}

\subsection{Classification using variational posterior}

Define, $\hat{\eta}(\boldx)$, the variational estimator of $\eta_0(\boldx)$ as
\begin{equation}
    \label{e:eta-hat-def}
\hat{\eta}(\boldx)=\sigma^{-1}\left(\int \sigma(\eta_{\btheta_n}(\boldx)) \pi^*(\btheta_n)d\btheta_n \right)
\end{equation}
where $\pi^*$ is the variational posterior. Analgous to \eqref{e:Bayes}, the classifier based on $\hat{\eta}(\boldx)$ is given by
\begin{equation}
\label{e:Bayes-hat}
    \hat{C}(\boldx)=\begin{cases} 
1, & \sigma(\hat{\eta}(\boldx))\geq 1/2\\
0, & \text{otherwise}
\end{cases}
\end{equation}
Note, the formulation in \eqref{e:eta-hat-def} guarantees that we directly approximate the main quantity of interest, $\sigma(\eta_0(\boldx))$ as in \eqref{e:class}  by its posterior mean, $\int  \sigma(\eta_{\btheta_n}(\boldx)) \pi^*(\btheta_n)d\btheta_n $, which is empirically estimated as
\begin{equation}
    \label{e:eta-hat-approx}
    \hat{\eta}^{M}(\boldx)=\frac{1}{M}\sum_{i=1}^M \sigma(\eta_{\btheta_n[i]}(\boldx))
\end{equation}
where $\btheta_n[1], \cdots, \btheta_n[M]$ are multiple samples from the variational posterior $\pi^*$. Since generation of multiple samples from the variational posterior is much cheaper, the order of error between \eqref{e:eta-hat-def} and \eqref{e:eta-hat-approx} is negligible.

\section{Theoretical Properties: Posterior and Classification Consistency} 
\label{Sec:Theory}
In this section, we establish that the Bayesian inference procedure  proposed in section \ref{sec:bayes} enjoys theoretical guarantees in terms of consistency of the posterior estimation and classification. For a simple Gaussian mean field family as in \eqref{e:VIF}, we establish that the variational posterior  \eqref{e:var-posterior} is consistent under suitable assumptions on the prior parameters. We also discuss  how the  the true function $\eta_0$ impacts the rate of consistency of the variational posterior. Finally, we  present how the consistency rates of the variational posterior differ from those of the true posterior.

Let $f_0$ and $f_{\btheta_n}$ be the joint density of the observations $(y_i, \boldx_i)_{i=1}^n$ under the truth and the model respectively. Next, without loss of generality, we assume  $X_i \sim U[0,1]^p$, which implies $f_0(\boldx)=1$ and $f_{\btheta_n}(\boldx)=1$. This further implies that the joint distribution of $(y_i,\boldx_i)_{i=1}^n$ under the model and the truth depend on the conditional distribution of $Y|X=\boldx$. In view of \eqref{e:class} and \eqref{e:sig-x},
\begin{eqnarray}
\nonumber   
   f_{\btheta_n}(y,\boldx)&=&f_{\btheta_n}(y|\boldx) f_{\btheta_n}(\boldx)\:\:=\:\:\exp\left(y\eta_{\btheta_n}(\boldx)-\log \left(1+e^{\eta_{\btheta_n}(\boldx)}\right)\right)\:\:=\:\:\ell_{\btheta_n}
    (y,\boldx)\\
    f_0(y,\boldx)&=&f_0(y|\boldx)f_0(\boldx)\:\:\:\:\:\:\:=\:\:\:\exp\left(y\eta_0(\boldx)-\log \left(1+e^{\eta_0(\boldx)}\right)\right)\:\:\:\:\:=\:\:\ell_0(y,\boldx)
\end{eqnarray}
where  $\ell_{\btheta_n}$ and $\ell_0$ are defined in  \eqref{e:lik-one} and \eqref{e:lik-one-0} respectively.

We next define the Hellinger neighborhood of the true function density function $f_0=\ell_0$ as
\begin{equation}
\label{e:hell-def}
    \mathcal{U}_\varepsilon=\{\btheta_n: d_{\rm H}(\ell_0,\ell_{\btheta_n})<\varepsilon\}
\end{equation}
where the Hellinger distance, $d_{\rm H}(\ell_0,\ell_{\btheta_n})$ is given by 
$$d_{\rm H}(\ell_0,\ell_{\btheta_n})=\left(\frac{1}{2}\int_{\boldx \in [0,1]^p} \sum_{y \in \{0,1\}} \left(\sqrt{\ell_0(y,\boldx)}-\sqrt{\ell_{\btheta_n}(y,\boldx)}\right)^2  d\boldx\right)^{1/2}.$$

We also define the Kullback-Leibler (KL) neighborhood of the true function density function $f_0=\ell_0$ as
\begin{equation}
\label{e:kl-nei-def}
    \mathcal{N}_\varepsilon=\{\btheta_n: d_{\rm KL}(\ell_0,\ell_{\btheta_n})<\varepsilon\}
\end{equation}
where the KL distance, $d_{\rm KL}(\ell_0,\ell_{\btheta_n})$ is given by
\begin{equation*}
\label{e:kl-def}
d_{\rm KL}(\ell_0,\ell_{\btheta_n})=\int_{\boldx \in [0,1]^p} \sum_{y \in \{0,1\}}\left( \log  \frac{\ell_0(y,\boldx)}{\ell_{\btheta_n}(y,\boldx)} \ell_0(y,\boldx)\right) d\boldx
\end{equation*}

\noindent We use the notation  $P_0^n$ to denote the true distribution of $(\boldy_n,\boldX_n)=(y_i,\boldx_i)_{i=1}^n$ under the true density $\ell_0$.

\subsection{Posterior consistency and its implication in practice}

In the following two theorems for two class of priors, we establish the posterior consistency of $\pi^*$ defined in \eqref{e:var-posterior}. In this direction, we show that the variational posterior concentrates in $\varepsilon-$ small Hellinger neighborhoods of the true density $\ell_0$. In theorem \ref{thm:post-cons}, we establish this result for a fixed choice of the neighborhood distance $\varepsilon$. In theorem \ref{thm:post-cons-del}, we establish the same result for shrinking neighborhood sizes of the true function $\ell_0$. For both these theorems, the number of nodes $k_n$ are allowed to grow at a rate of $n^a$ for some $0<a<1$. However, since the theorem \ref{thm:post-cons-del} is more restrictive in nature, it requires certain assumptions on the approximating neural network solution to the true function $\eta_0$ (see assumption (A3) below). 
 Note that the theorem \ref{thm:post-cons} is a weaker convergence result, however, it is free from assumptions on the approximating neural network solution. theorem \ref{thm:post-cons-del} on the other hand requires the existence of a neural network solution  which converges to the true function $\eta_0$ at a fast enough rate while ensuring controlled growth of the $L_2$ norm of its coefficients. Additionally, the rate of growth of $L_2$ norm of the prior mean parameter is allowed to grow faster in theorem \ref{thm:post-cons} compared to theorem \ref{thm:post-cons-del} (see assumptions (A2) and (A4) below).

\begin{theorem}
\label{thm:post-cons}
Suppose  $k_n \sim n^a$, $0<a<1$. Additionally,

\noindent {\bf (A2)} The prior parameters in \eqref{e:prior} satisfies assumption \text{(A1)} and 
\begin{eqnarray*}
||\bmu_n||^2_2&=&o(n)
\end{eqnarray*}
Then, 
$$\pi^*(\mathcal{U}_{\varepsilon}^c) \stackrel {P_0^n}{\longrightarrow} 0$$
\end{theorem}
Here, $||.||_2$ is the $L_2$ norm of a vector as in definition \ref{def:norms}.

A restatement of the above theorem is for any $\nu>0$, $\pi^*(\mathcal{U}_{\varepsilon}^c)<\nu$ with probability tending to 1 as $n \to \infty$. Under the conditions of theorem \ref{thm:post-cons}, it can be established that the true posterior satisfies $\pi(\mathcal{U}_{\varepsilon}^c|\boldy_n,\boldX_n)<2e^{-n\varepsilon^2/2}$ with probability tending to 1 as $n \to \infty$ (see theorem  \ref{thm:true-post-cons} part 1. in appendix (C) of section \ref{sec:append}). This implies that the probability of the $\varepsilon-$Hellinger neighborhoods of the true function $\ell_0$ for the true posterior increases at the rate of $1-2e^{-n\varepsilon^2/2}$ in contrast to the slow rate of $1-\nu$ for the variational posterior.

\begin{theorem}
\label{thm:post-cons-del}
Suppose  $k_n \sim n^a$, $0<a<1$ and $\epsilon_n^2 \sim n^{-\delta}$, $0<\delta<1-a$. Additionally,

\noindent{\bf (A3)} There exists a sequence of neural network functions $\eta_{\boldt_n}$ of the form \eqref{e:eta-x} satisfying 
\begin{eqnarray*}
||\eta_0-\eta_{\boldt_n}||_\infty&=&o(\epsilon_n^2)\\
||\boldt_n||_2^2&=&o(n\epsilon_n^2)  
\end{eqnarray*}
{\bf (A4)} The prior parameters in \eqref{e:prior} satisfies assumption (A1) and 
\begin{eqnarray*}
||\bmu_n||^2_2&=&o(n\epsilon_n^2)
\end{eqnarray*}
Then, 
$$\pi^*(\mathcal{U}_{\varepsilon \epsilon_n}^c) \stackrel {P_0^n}{\longrightarrow} 0$$
\end{theorem}
Here, $||.||_2$ is the $L_2$ norm of a vector and $||.||_\infty$ is the $L_\infty$ norm of a function as in \ref{def:norms}.

A restatement of the above theorem is for any $\nu>0$, $\pi^*(\mathcal{U}_{\varepsilon \epsilon_n}^c)<\nu$ with probability tending to 1 as $n \to \infty$. Under the conditions of theorem \ref{thm:post-cons-del}, it can be established that the true posterior satisfies $\pi(\mathcal{U}_{\varepsilon  \epsilon_n}^c|\boldy_n,\boldX_n)<2e^{-n  \varepsilon^2\epsilon_n^2/2}$ with probability tending to 1 as $n \to \infty$ (see theorem  \ref{thm:true-post-cons-1} part 1. in appendix (C) of section \ref{sec:append}). This implies that the probability of the shrinking  $\varepsilon \epsilon_n-$Hellinger neighborhoods of the true function $\ell_0$ for the true posterior increases at the rate of $1-2e^{-n \varepsilon^2\epsilon_n^2/2}$ in contrast to the slow rate of $1-\nu$ for the variational posterior.

\subsection{Discussion of the proof} 

We next briefly outline the main steps in the the proof of theorems \ref{thm:post-cons} and \ref{thm:post-cons-del}. The details are deferred to appendix (B) in section \ref{sec:append}. We borrow a few steps and notations from \cite{BHAT2020}.  The first step of the proof is to establish that $d_{\rm KL}(\pi^*, \pi(.|{\boldy}_n,{\boldX}_n))$ is bounded below by a quantity which is determined by the rate of consistency of the true posterior. The second step towards the proof is to show  $d_{\rm KL}(\pi^*, \pi(.|{\boldy}_n,{\boldX}_n))$ is bounded above at a rate which is greater than its lower bound if and only if the variation posterior is consistent.
In this direction, with $\mathcal{U}_\varepsilon$ as in \eqref{e:hell-def} note that for any $\varepsilon>0$

\begin{eqnarray*}
\label{e:dk-b}
&&\hspace{-15mm}d_{\rm KL}(\pi^*, \pi(.|{\boldy}_n,{\boldX}_n))\\
&=&\int_{\mathcal{U}_\varepsilon} \pi^*(\btheta_n)\log \frac{\pi^*(\btheta_n)}{\pi(\btheta_n|\boldy_n,\boldX_n)}d\btheta_n+\int_{\mathcal{U}_\varepsilon^c} \pi^*(\btheta_n)\log \frac{\pi^*(\btheta_n)}{\pi(\btheta_n|\boldy_n,\boldX_n)}d\btheta_n\\
&=&-\pi^*(\mathcal{U}_\varepsilon)\int_{\mathcal{U}_\varepsilon} \frac{\pi^*(\btheta_n)}{\pi^*(\mathcal{U}_\varepsilon)}\log \frac{\pi(\btheta_n|\boldy_n,\boldX_n)}{\pi^*(\btheta_n)}d\btheta_n-\pi^*(\mathcal{U}_\varepsilon^c)\int_{\mathcal{U}_\varepsilon^c} \frac{\pi^*(\btheta_n)}{\pi^*(\mathcal{U}_\varepsilon^c)}\log \frac{\pi(\btheta_n|\boldy_n,\boldX_n)}{\pi^*(\btheta_n)}d\btheta_n\\
&&\\
&\geq& \pi^*(\mathcal{U}_\varepsilon) \log 
	\frac{\pi^*(\mathcal{U}_\varepsilon)}{\pi(\mathcal{U}_\varepsilon|\boldy_n,\boldX_n)}+\pi^*(\mathcal{U}_\varepsilon^c) \log 
	\frac{\pi^*(\mathcal{U}_\varepsilon^c)}{\pi(\mathcal{U}_\varepsilon^c|\boldy_n,\boldX_n)}, \hspace{2mm}\text{by Jensen's inequality}
	\end{eqnarray*}
Since $\pi(\mathcal{U}_\varepsilon|\boldy_n,\boldX_n)\leq 1$, thus
\begin{eqnarray*}
\hspace{5mm}&\geq& \pi^*(\mathcal{U}_\varepsilon) \log 
	\pi^*(\mathcal{U}_\varepsilon)+\pi^*(\mathcal{U}_\varepsilon^c) \log 
	\pi^*(\mathcal{U}_\varepsilon^c)-\pi^*(\mathcal{U}_\varepsilon^c) \log \pi(\mathcal{U}_\varepsilon^c|\boldy_n,\boldX_n)\\
	&\geq& -\pi^*(\mathcal{U}_\varepsilon^c) \log \pi(\mathcal{U}_\varepsilon^c|\boldy_n,\boldX_n)-\log 2, 
	\hspace{4mm}\text{since }x\log x+(1-x)\log (1-x) \geq -\log 2
	\\
	&=&-\pi^*(\mathcal{U}_\varepsilon^c) \left(\log \int_{\mathcal{U}_\varepsilon^c} \frac{L(\btheta_n)}{L_0}p(\btheta_n)d\btheta_n-\log \int \frac{L(\btheta_n)}{L_0}p(\btheta_n)d\btheta_n\right)-\log 2
\end{eqnarray*}
Thus, with
\begin{equation}
\label{e:A-B-def}
      A_n=\log \int_{\mathcal{U}_\varepsilon^c} \frac{L(\btheta_n)}{L_0}p(\btheta_n)d\btheta_n \hspace{10mm}
    B_n=-\log \int \frac{L(\btheta_n)}{L_0}p(\btheta_n)
d\btheta_n 
\end{equation}
we get the following main step towards the proof of theorems \ref{thm:post-cons} and \ref{thm:post-cons-del}.  
\begin{equation}
\label{e:kl-ub-lb}
 \boxed{-\pi^*(\mathcal{U}_\varepsilon^c) A_n 
 \leq d_{\rm KL} (\pi^*, \pi(.|\boldy_n,\boldX_n))+|B_n|+\log 2}
\end{equation}

In the above proof we have assumed $\pi^*(\mathcal{U}_\varepsilon)>0$, $\pi^*(\mathcal{U}_\varepsilon^c)>0$. If $\pi^*(\mathcal{U}_\varepsilon^c)=0$, there is nothing to prove. If $\pi^*(\mathcal{U}_\varepsilon)=0$, then following the steps of the proof in appendix (B) in section \ref{sec:append}, we will get $\varepsilon^2=o_{P_0^n}(1)$ which is a contradiction. 

The exponential of the first term $A_n$ is decomposed as $$e^{A_n}=\int_{\mathcal{U}_\varepsilon^c \cap \mathcal{F}_n} \frac{L(\btheta_n)}{L_0}p(\btheta_n)d\btheta_n+\int_{\mathcal{U}_\varepsilon^c \cap \mathcal{F}_n^c}   \frac{L(\btheta_n)}{L_0} p(\btheta_n)d\btheta_n$$
where ${\{\mathcal{F}_n\}}_{n=1}^\infty$ is a suitably chosen sequence of sieves. Indeed our choice of $\mathcal{F}_n$ is given by
\begin{equation}
\label{e:Fn-def}
\mathcal{F}_n=\Big\{\btheta_n:|\theta_{jn}|\leq C_n, j=1,\cdots, K(n)\Big\}
\end{equation}
where $C_n=e^{n^b/K(n)}$ in theorem \ref{thm:post-cons} and $C_n=e^{n^b\epsilon_n^2/K(n)}$ in theorem \ref{thm:post-cons-del} respectively. The constant $b$ is chosen suitably to ensure Hellinger bracketing entropy (see definition \ref{def:hell-brack}) of $\mathcal{F}_n$ is well controlled (see lemma \ref{lem:v-bound} for more details). Secondly, the prior needs to give negligible probability outside $\mathcal{F}_n^c$  so that term $e^{A_n}$ is well controlled. The  prior in \eqref{e:prior}  satisfies this criterion for theorem \ref{thm:post-cons} and theorem \ref{thm:post-cons-del} with assumptions (A1), (A2) and (A1), (A4) respectively.

The second quantity $B_n$ is controlled by the rate at which the prior gives mass to shrinking KL neighborhoods of the true density $\ell_0$. In theorem \ref{thm:post-cons}, this rate is controlled as long as the prior parameter of the prior in \eqref{e:prior} satisfies (A1) and (A2). In theorem \ref{thm:post-cons-del}, the same rate is controlled as long as the prior parameters satisfies (A1) and (A4) and the  true function $\eta_0$ has a neural network solution which satisfies assumption (A3).

Finally, we  bound $d_{\text{KL}} (\pi^*, \pi(.|\boldy_n,\boldX_n))$ by $d_{\text{KL}} (q, \pi(.|\boldy_n,\boldX_n))$ for a suitable $q\in \mathcal{Q}_n$ (see proposition \ref{lem:f-bound} and proposition \ref{lem:q-bound-v} for more details). From \eqref{e:d-KL-break}, note that
\begin{equation}
\label{e:dkl-n}
d_{KL} (q, \pi(.|\boldy_n,\boldX_n)) \leq d_{\rm KL}(q,p)+\left|\int \log  \frac{L(\btheta_n)}{L_0}q(\btheta_n) d\btheta_n\right|+\left|\log \int \frac{L(\btheta_n)}{L_0}  p(\btheta_n) d\btheta_n\right|
\end{equation}
In the above expression, the last term is nothing but $|B_n|$. The second term is the most crucial quantity of interest.
$$    \left|\int \log  \frac{L(\btheta_n)}{L_0}q(\btheta_n) d\btheta_n\right|\approx n \int d_{\text{KL}}(\ell_0,\ell_{\btheta_n}) q(\btheta_n) d\btheta_n.$$
For both the theorems \ref{thm:post-cons} and \ref{thm:post-cons-del}, the right hand side can always be controlled by choosing $q=MVN(\boldt_n,I_{K(n)}/\sqrt{n})$ for a suitable choice of the sequence $\boldt_n$. For theorem  \ref{thm:post-cons}, this sequence corresponds to  $\eta_{\boldt_n}$, the finite neural network approximation of $\eta_0$ and for theorem  \ref{thm:post-cons-del}, this sequence corresponds  to $\eta_{\boldt_n}$, the rate controlled neural network approximation of assumption (A3). Finally, the first term in \eqref{e:dkl-n} is determined by both prior and $q$. In theorem \ref{thm:post-cons}, it is controlled as long as the prior parameter of the prior in \eqref{e:prior} satisfies (A1), (A2). In theorem \ref{thm:post-cons-del}, the same rate is controlled as long as the prior parameters satisfies (A1), (A4) and the sequence $\boldt_n$ satisfies assumption (A3) part 2.

In light of the above discussion, there are three main properties which a prior must satisfy to allow for the convergence of variational posterior. For any $\nu>0$
\begin{enumerate}
    \item For a sequence of sieves ${\{\mathcal{F}_n\}}_{n=1}^\infty$ with well controlled Hellinger bracketing entropy, $$\int_{\mathcal{F}_n^c} p(\btheta_n)d\btheta_n \leq e^{-n \nu}, n \to \infty$$
        \item With $\mathcal{N}_\varepsilon$ as in \eqref{e:kl-nei-def},
        $$\int_{\mathcal{N}_\varepsilon} p(\btheta_n)d\btheta_n \geq e^{-n \nu}, n \to \infty$$
    \item For a $q$ satisfying $\int d_{\text{KL}}(\ell_0,\ell_{\btheta_n})q(\btheta_n)d\btheta_n<\varepsilon, n \to \infty$,
    $$d_{KL}(q,p) \leq n\nu, n \to \infty$$
\end{enumerate}
Whereas the condition 1 and 2, are standard assumptions for consistency of true posterior (see assumptions 1 and 2  in \cite{BSW} and theorem 2 in \cite{LEE}), condition 3 is an additional requirement which makes the variational posterior consistent. Indeed, the proof presented in this section can be generalized to a much wider class of priors satisfying conditions (1)-(3).

\subsection{Classification consistency}

In this section, we discuss the classification accuracy of the predictions made by the  variational posterior by comparing to the optimal mis-classification error. In view of \eqref{e:riskintegral}, let $R(\hat{C})$ and $R(C^{\text{Bayes}})$ denote the classification accuracy under the variational classifier in \eqref{e:Bayes-hat} and the Bayes classifier in \eqref{e:Bayes} respectively. We next establish the how the difference in classification accuracy depends on the logit links $\hat{\eta}(X)$ and $\eta_0(X)$ as defined in \eqref{e:eta-hat-def} and \eqref{e:class} respectively.
\begin{align}
 \nonumber   &R(\hat{C})-R(C^{\rm Bayes})=E_{X}E_{Y|X}[I_{\hat{C}(X) \neq Y}-I_{C^{\rm Bayes}(X) \neq Y}]\\
\nonumber    &=E_{X} E_{Y|X}[(I_{\hat{C}(X) =0} -I_{C^{\rm Bayes}(X) =0}) \sigma(\eta_0(X))+(I_{\hat{C}(X) =1}-I_{C^{\rm Bayes}(X) =1})(1-\sigma(\eta_0(X)))]\\
\nonumber    &=2E_{X}[I_{\hat{C}(X)\neq C^{\rm Bayes}(X)}|\sigma(\eta_0(X))-1/2|]\\
\nonumber    &=2E_{X}[I_{\sigma(\hat{\eta}(X))\geq 1/2,\sigma(\eta_0(X))< 1/2}|\sigma(\eta_0(X))-1/2|+I_{\sigma(\hat{\eta}(X))< 1/2,\sigma(\eta_0(X))\geq  1/2}|\sigma(\eta_0(X))-1/2|]\\
    &\leq 2E_X|\sigma(\eta_0(X))-\sigma(\hat{\eta}(X))|
    \label{e:r-c-eta}
\end{align}

Using the above result, in corollary  \ref{cor:fun-cons}, we establish the classification accuracy of the variational estimate $\hat{\eta}(\boldx)$ under no assumptions on the true function $\eta_0(\boldx)$. In corollary \ref{cor:fun-cons-del}, we establish the same result under assumption (A3) on the true function $\eta_0(\boldx)$. Note, although theorem \ref{thm:post-cons} requires  minimal assumptions, it gives a much weaker convergence result on the classification accuracy.

\begin{corollary}
\label{cor:fun-cons}
Under the conditions of theorem \ref{thm:post-cons},
$$|R(\hat{C})-R(C^{\rm Bayes})| \stackrel{P_0^n}{\longrightarrow}0$$
\end{corollary}

A restatement of the above corollary is for any $\nu>0$, $|R(\hat{C})-R(C^{\rm Bayes})|<\nu$ with probability tending to 1 as $n \to \infty$. Under the conditions of theorem \ref{thm:post-cons}, it can be established that the true posterior also gives classification consistency at the same rate and there is no loss in using a variational posterior approximation (see theorem \ref{thm:true-post-cons} part 2. in appendix (C) of \ref{sec:append}).

\begin{corollary}
\label{cor:fun-cons-del}
Under conditions of theorem \ref{thm:post-cons-del}, for every $0\leq \kappa\leq  2/3$,
$$\epsilon_n^{-\kappa}|R(\hat{C})-R(C^{\rm Bayes})| \stackrel{P_0^n}{\longrightarrow}0$$
\end{corollary}

A restatement of the above corollary is for any $\nu>0$, $0\leq \kappa \leq 2/3$,  $|R(\hat{C})-R(C^{\rm Bayes})|<\nu \epsilon_n^{k}$ with probability tending to 1 as $n \to \infty$. Under the conditions of theorem \ref{thm:post-cons-del}, it can be established that the true posterior satisfies  $ |R(\hat{C})-R(C^{\rm Bayes})|<\nu \epsilon_n^{k}$ for every $\nu>0$, $0\leq\kappa\leq 1 $ with probability tending to 1 as $n \to \infty$  (see theorem \ref{thm:true-post-cons-1} part 2. in appendix (C) of \ref{sec:append}). Thus, the classification consistency occurs at the rate $\epsilon_n^{2/3}$ for the variational posterior in contrast to $\epsilon_n$ for the true posterior.

\section{Numerical Properties and Alzheimer’s Disease Study}
\label{Sec:Numerical}

The transition from mild cognitive impairment (MCI) to Alzheimer’s disease (AD) is of great interest for clinical researchers. 
Several studies over the past decade have shown and compared the performance of different machine learning methods on this classification task. 
 For this classification problem, we illustrate the performance of variational Bayesian neural networks as developed under section \ref{sec:bayes} in terms of classification accuracy, numerical complexity and time of convergence. We implemented both algorithms, algorithm \ref{al:bbvi} and \ref{al:bbvi-cv} and shall hence forth refer to them as BBVI and BBVI-CV respectively. For a comparative baseline, we also report the performance for several machine learning techniques as applicable to this task. We like to emphasize that, our primary goal here is to illustrate the computational methodology rather incremental improvement for a specific application. 
 


Alzheimer’s disease (AD) is a progressive, age-related, neurodegenerative disease and the most common cause of dementia \cite{zhang,zhang2012,Koro}. Behaviorally, AD is commonly preceded by mild cognitive impairment (MCI), a syndrome characterized by decline in memory and other cognitive domains that exceed cognitive decrements associated with normal aging \cite{zhang2012,Petersen}. However, the prodromal symptoms of MCI are not prognostically deterministic: individuals with MCI tend to progress to probable AD at a rate of $8\%$-$15\%$ per year, and most conversions occur within 3 years of presentation  \cite{cui2011,Far,Allision}. Research efforts to provide new insights into the incidence of MCI-to-AD conversion have focused largely on clinically or biologically relevant features (i.e., neuroimaging markers, clinical exam data, neuropsychological test scores) and on different methods for statistical classification \cite{Young}.
We used T1-weighted MRI images from the collection of standardized datasets. The description of the standardized MRI imaging from ADNI can be found in http://adni.loni.usc.edu/methods/mri-analysis/adni-standardized-data.

This study used a subset of the MCI subjects from ADNI-1, who had data from demographic, clinical cognitive assessments, APOE4 genotyping, and MRI measurements. In total, there are 819 individuals with a baseline diagnosis of MCI, but we only consider patients whose follow-up period was at least 36 months and no missing values.
The final samples included 265 subjects which included participants who were stable in their diagnosis (MCI-S) and those who converted to a diagnosis of AD over 3 years (MCI-C).
We considered a total of 18 clinical as potential predictors of MCI-to-AD progression in our classification analyses. These included scores on the Mini Mental State Examination (MMSE), Clinical Dementia Rating Sum of Boxes (CDR-SB), Alzheimer’s Disease Assessment Scale-cognitive subscale (ADAS-cog), Activities of Daily living (from the Functional Activities Questionnaire, FAQ), Trail Making tests B (TRABSCOR), The Rey Auditory Verbal Learning Test (RAVLT), The Digit- Symbol Coding test (DIGT). We also considered genotype for carriers of the epsilon-4 allele of the apolipoprotein E (APOE) gene \cite{Young} as a genetic predictor in this study. Table \ref{table:clinical_predictor} in appendix (D) of section \ref{sec:append} summarizes all 18 clinical, demographic and genetic features used in this study. Structural MRI data were collected according to the ADNI acquisition protocol using T1-weighted scans (GradWarp, B1 Correction, N3, Scaled). This was followed by brain extraction for further processing \cite{Doshi2013}. A new multi-atlas registration based label fusion method  was applied for region of interest (ROI) segmentation \cite{Doshi2016}. These data include baseline MRI scans of ADNI-1 participants. MRI scans were automatically partitioned into 145 anatomic ROIs spanning the entire brain. An additional 114 derived ROIs were calculated by combining single ROIs within a tree hierarchy, to obtain volumetric measurements from larger structures \cite{Doshi}. In total, 259 ROIs were measured and used as potential predictors of MCI-to-dementia progression in this study. Based on the extant literature \cite{korolev9-13,zihuanL}, we used 24 ROI features as theoretically significant of MCI to dementia progression (see table \ref{table:roi} in appendix (D) of section \ref{sec:append}). 

Note that the dependence and interactions among different modes of features (clinical, MRI) and within the modes may be different and hard to model explicitly. Thus a neural network-based modeling is intuitive from predictive modeling and machine learning perspective. Out of the 265 patients, every time 186 are selected by simple random sample as training cases and the remaining 79 as test cases. The approximate 2:1 ratio for training and test cases is, of course, arbitrary. All the covariates (except categorical variables) were z-normalized.  The outcome $y_i$ for the $i$th patient is either 1 for MCI-C or 0 for MCI-S in classification study. We adopted a 10-fold cross-validation to avoid optimistically-biased estimates of model performance.

\subsection*{Theory guided parameter choice for statistical and computational models.}
In order to implement the BBVI and BBVI-CV, we first need to make a valid choice of the prior parameters $\mu_{jn}$, $\zeta_{jn}$ for $j=1, \cdots, K(n)$ as in \eqref{e:prior}. We use the choice of $\mu_{jn}=0$ and $\zeta_{jn}=1$ for our prior parameters. Indeed, this choice satisfies conditions (A1), (A2) and (A4) as assumed in the consistency proofs of theorems \ref{thm:post-cons} and \ref{thm:post-cons-del}. Next, for the implementation of the BBVI and BBVI-CV, we need to make a choice on the number nodes $k_n$. We tried with $k_n=2,10,20$ and obtained the best results at $k_n=10$, the results of which are reported in this paper. Note that $k_n=10$ satisfies the assumption of theorem \ref{thm:post-cons} and \ref{thm:post-cons-del} with $a=0.44$. The next two important parameters of  interest in the implementation of  BBVI and BBVI-CV are the sample size $S$, used in computation of the Monte Carlo estimate of the gradient \eqref{e:mcmcelbo-0} and  the learning rate $\rho_t$ used in the updation rule \eqref{e:grad-update}. We next discuss the performance of algorithms \ref{al:bbvi} and \ref{al:bbvi-cv} in light of these two parameters. 

\subsection*{Choice of the learning rate} 
The stochastic optimization task is to find variational parameter vector $\mathcal{V}_q$ which maximizes the evidence lower bound ELBO in  \eqref{e:elbo-idx} or consequently minimizes the Kullback-Leibler (KL) distance in \eqref{e:KL_qp}. Even though the BBVI is straightforward in its general definition, the choice of learning rate $\rho_t$ can be challenging in practice. Ideally, one would want the rate to be small in situations where the noisy estimates of the gradient have large variance and vice-versa. The elements of variational parameter can also differ in scale, and one needs to set the learning rate so that both the BBVI and BBVI-CV can accommodate even the smallest scales. So, the first issue is addressed in this paper by choosing the learning rate $\rho_t$ such that we can obtain fast convergence rate of the ELBO  to a good local maximum.  The $\rho_t$ considered in this paper are the following: 1) constant $\rho_t=\rho$; 2) varying learning rate with $\rho_t=\frac{\rho_0}{b(t+1)^{c}}$, where $b>0$ and  $c>0$. On  one hand, a small constant rate generally does not allow the parameter vector to converge, a large constant rate can lead to slower convergence.  Indeed, we would like a learning rate which guarantees learning algorithms converge, and converge as quickly as possible. For varying learning rate, $\rho_t$ should satisfy the Robbins and Monro conditions: $\sum_t \rho_t=\infty$ and $\sum_t \rho_t^2 < \infty$ to make BBVI and BBVI-CV converge to a local maximum \cite{Christian91,RobbinsH}. In this work, we found using a fixed learning rate with $\rho_t=0.0001$ and a varying learning rate with $\rho_0=1$, $b=100$ and $c=0.3$ works well.

 
\subsection*{Choice of the sample size $S$} 
The choice of sample size $S$ is sensitive to the performance to model in terms of algorithmic stability and convergence time. Whereas each update with small sample size takes less time,  the variability of the estimate is high. On the other hand a large sample size leads to less variable estimates but each update takes a much longer time. We would thus like to have a sample size which is just large enough so that the learning algorithms can offer best testing accuracy and faster convergence rate at the same time.  We started with relatively small $S$ for learning the robustness of machines, then we increased the sample size until we obtained the best model's performance both in terms of stability of variance of gradients and testing accuracy. Specifically, we experimented with $S=200$, $S=500$ and $S=1000$. 

\subsection*{Variance of gradient estimators} 
 As explained in section \ref{sec:cv}, the maximization of the ELBO in \eqref{e:elbo} requires stabilization of the variance of the stochastic gradient in \eqref{e:mcmcelbo-0}. Indeed there are two approaches which can effectively reduce the variance of the gradients (1) increasing the sample size $S$ in Monte Carlo estimation of the gradient and (2) using a control variate approach which is described in section \ref{sec:cv}. In this direction, we try to choose the balance between these two variance reduction techniques. Figure \ref{plot:var of gradients,models} illustrates how the variance of gradient estimators changes with the number of iterations for a given $S$. 
 Figure \ref{plot:var of gradients,samples} illustrates the difference in variance of gradient estimators as $S$ changes from 200 to 1000. For both these figures, the variance of gradient in \eqref{e:mcmcelbo-0} is estimated by considering the average of the empirical variance calculated across all its $\mathcal{V}_q$ components.
 Note that the gradients are more stable using BBVI-CV with smaller  difference as we go from $S=200$ to $S=1000$. Further the variance under BBVI-CV is smaller comapered to BBVI.
This investigation suggests that applying control variates and using larger sample size reduces the variance significantly and this is expected. Thus, the optimal choice is decided by the combination which produces fast convergence and better test accuracy.

\begin{figure}[!h]
\centering
\begin{subfigure}{0.5\linewidth}
  \centering
  \includegraphics[width=\linewidth]{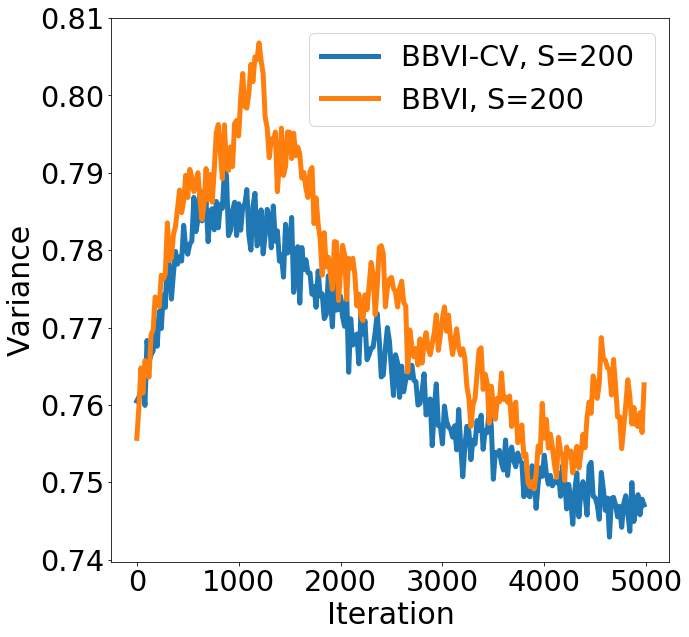}
  \label{}
\end{subfigure}%
\begin{subfigure}{0.5\linewidth}
  \centering
  \includegraphics[width=\linewidth]{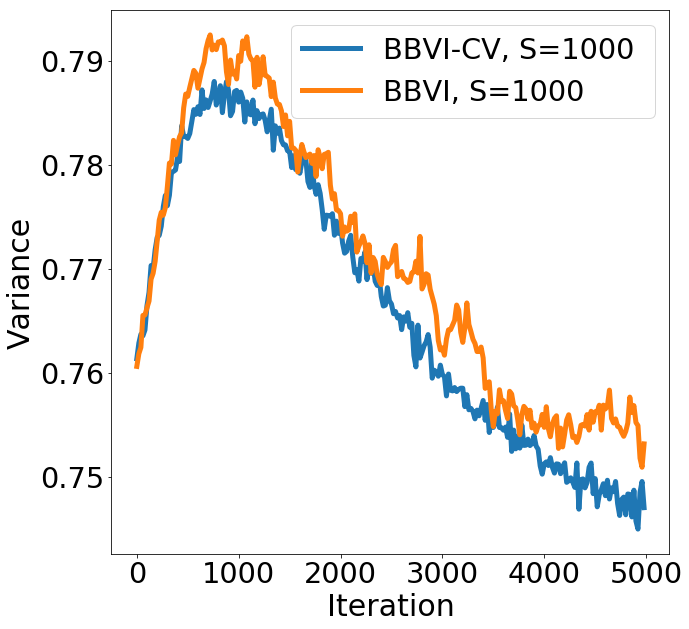}
  \label{}
\end{subfigure}
\caption{
 Variance of gradients for varying algorithms and fixed $S$.}
\label{plot:var of gradients,models}
\end{figure}

\begin{figure}[!h]
\centering
\begin{subfigure}{0.5\linewidth}
  \centering
  \includegraphics[width=\linewidth]{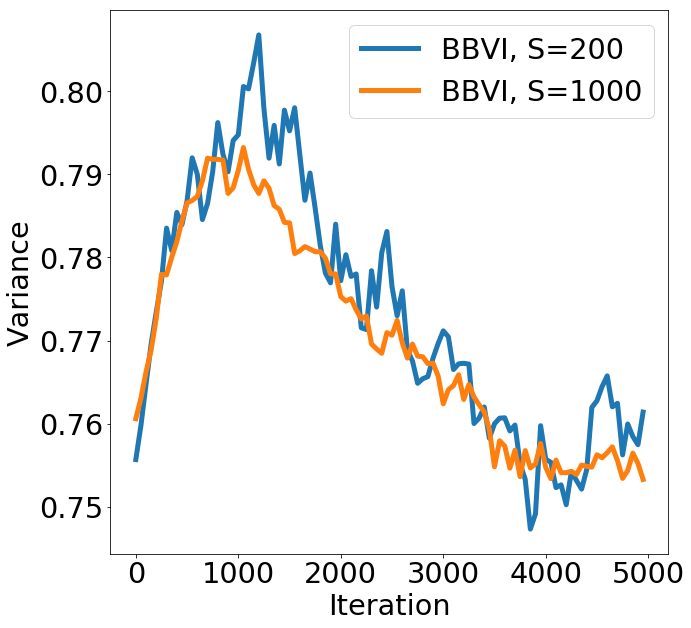}
  \label{}
\end{subfigure}%
\begin{subfigure}{0.5\linewidth}
  \centering
  \includegraphics[width=\linewidth]{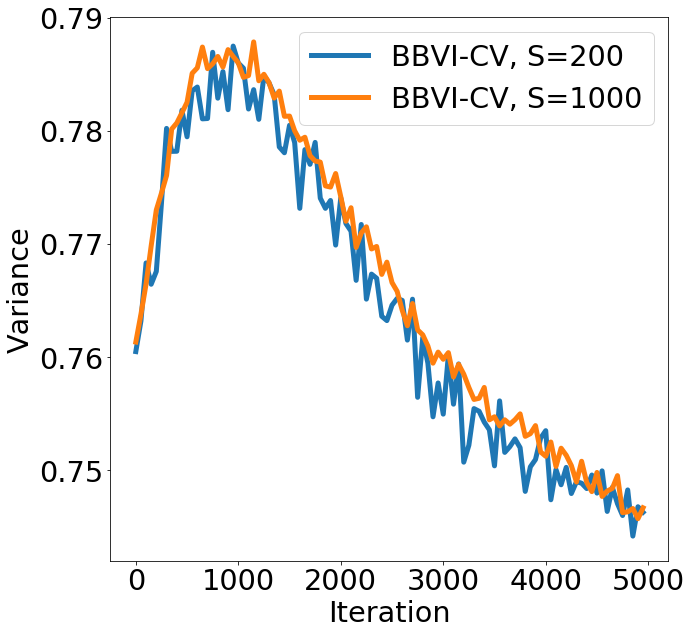}
  \label{}
\end{subfigure}
\caption{
Variance of gradients for varying $S$ and fixed choice of the algorithm. }
\label{plot:var of gradients,samples}
\end{figure}

\subsection*{Testing accuracy and convergence.}
We evaluated the model's performance for both the algorithms,  BBVI and BBVI-CV under  two criteria (1) testing accuracy (2) convergence time. The test accuracy of a classi:fier is given by
\begin{equation}
    \label{e:acc}
T(C) =\int_{\mathbb{R}^p \times \{0,1\}} I_{\{C(X,Y)=Y\}}dP_{X,Y} =1- R(C) 
\end{equation}
where $T(C)$ is the mis-classification error rate as described in \eqref{e:riskintegral}. The convergence criterion is defined as the point where Monte Carlo estimate of the ELBO as in \eqref{e:elbo-val} converges. We report the results with    several combinations of hyper-parameter in table \ref{Table:BBVI} : (a) $S=200$, $\rho_t=0.001$; (b) $S=500$, $\rho_t=0.001$; (c) $S=1000$, $\rho_t=0.001$; (d) $S=200$, $\rho_t$($b=100 \:,\: c=0.3$); (e) $S=500$,  $\rho_t$ ($b=100,\: c=0.3$); (f) $S=1000$, $\rho_t$ ($b=100,\: c=0.3$). 
In terms of the testing accuracy, the best model is BBVI with $S=1000$ and a fixed learning rate. This produces average testing accuracy  $75.89\%$ with standard error  $0.56\%$. 
The Table \ref{Table:BBVI} also provides convergence time (second) for one complete run of BBVI and BBVI-CV for one data split on a 2.3 GHz 8-Core Intel Core i9 MacBook Pro workstation. In terms of convergence time, the best model is BBVI-CV with $S=200$ for fixed learning rate.

\begin{table}[!h]
\small
\centering
\label{Result_t}
\scalebox{1.25}{
\begin{tabular}{@{}lccccc@{}} 
\toprule 
&  & \multicolumn{4}{c}{Testing accuracy(\%)}{Convergence time(s)}\\\cmidrule{3-6} 
Method & Sample size (S) & Fixed  & Variate& Fixed & Variate \\ \midrule 
BBVI & 200  & $74.46\pm 1.74$ & $74.11\pm 1.28$ & 1247 & 1860\\ 
& 500 & $74.28\pm 0.91$ & $74.64\pm 0.56$ & 1359 & 2061\\ 
& 1000 & $75.89\pm 0.56$ & $75.12\pm 0.80$&1652 & 2194\\ 
BBVI-CV & 200 & $75.71\pm0. 61$ &  $73.93\pm 5.95 $ & 1201 &  1268\\ 
 & 500 & $75.35\pm 0.56$ &  $74.11\pm 5.95$& 1263 &  1255\\ 
& 1000 & $75.00 \pm0.00$ &  $74.28\pm6.00$& 1735 &  1466\\\bottomrule 
\end{tabular}
}
 \caption{Performance for BBVI and BBVI-CV.} \label{Table:BBVI}
\end{table}

\begin{figure}[!h]
\centering
\begin{subfigure}{0.5\linewidth}
  \centering
  \includegraphics[width=\linewidth]{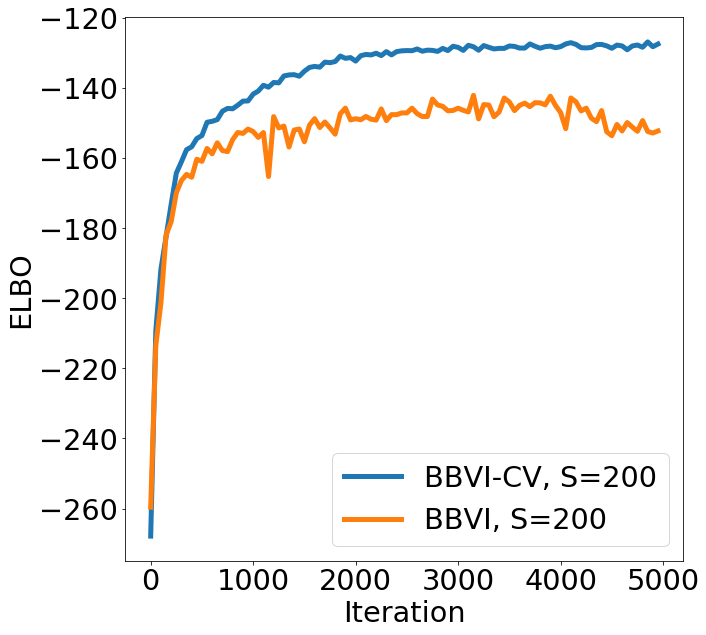}
  \label{}
\end{subfigure}%
\begin{subfigure}{0.5\linewidth}
  \centering
  \includegraphics[width=\linewidth]{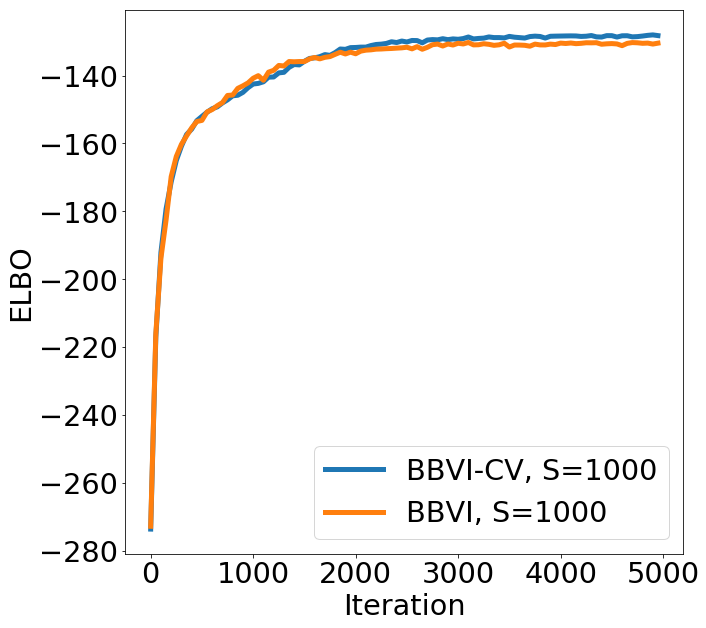}
  \label{}
\end{subfigure}
\caption{Convergence of the ELBO for varying algorithms and fixed $S$.}
\label{Fig:elbo,models}
\end{figure}

\begin{figure}[!h]
\centering
\begin{subfigure}{0.5\linewidth}
  \centering
  \includegraphics[width=\linewidth]{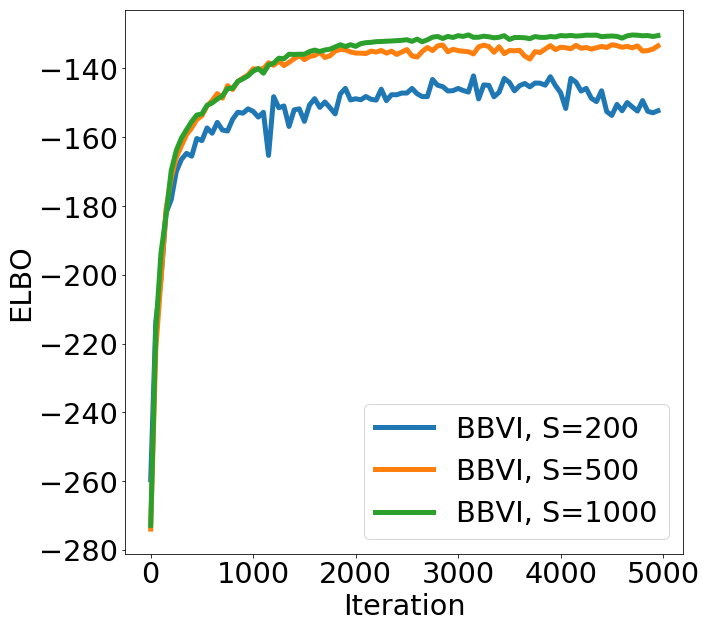}
  \label{}
\end{subfigure}%
\begin{subfigure}{0.5\linewidth}
  \centering
  \includegraphics[width=\linewidth]{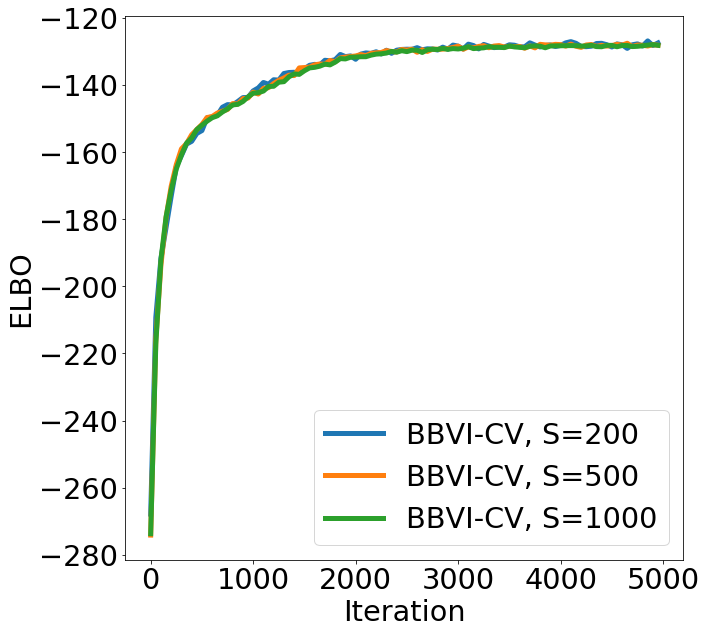}
  \label{}
\end{subfigure}
\caption{Convergence of ELBO for varying $S$ and fixed choice of the algorithm.}
\label{Fig:elbo,samples}
\end{figure}

For fixed learning rate, figure \ref{Fig:elbo,models} shows differences in  ELBO values between BBVI and BBVI-CV  for $S=200$ and $S=1000$ respectively in left and right panel. Figure \ref{Fig:elbo,samples} shows change in ELBO values  for $S=200,500,1000$ for BBVI and BBVI-CV respectively in left and right panel. Figure \ref{Fig:elbo,models} suggests ELBO difference in BBVI and BBVI-CV is negligible for higher value of $S$. Figure \ref{Fig:elbo,samples} indicates ELBO for BBVI-CV is less sensitive to the choice of $S$ compared to BBVI. Further, one can reach lower ELBO values much faster with BBVI-CV compared to BBVI.  BBVI model converges to same level of ELBO values after 4000 iterations where as BBVI-CV  took less than 2000 iterations to achieve same level of ELBO values. 
The convergence of ELBO shares same behavior as the stabilization of the variance of the gradient. Although at large values of $S$, convergence of ELBO needs smaller number of iterations,  each iteration itself takes much more time with large  $S$. From table \ref{Table:BBVI}, we  see that the overall convergence time is smaller with small values of $S$.  However, the testing accuracy is better at larger sample sizes. The  convergence time for BBVI-CV algorithm is overall smaller with comparable testing accuracy even at small values of $S$. 

Based on all these observations, we recommend the use of BBVI-CV with $S=200$ to allow good testing accuracy and smaller computation cost.


\subsection*{Numerical comparison with popular models} 
In this section, we numerically compare the testing accuracy of BBVI and BBVI-CV to a few benchmark models which include logistic regression (LR) and support vector machine (SVM) as developed by \cite{log_svm,python} and frequentist artificial neural network (ANN) \cite{chen2015mxnet,gur}.
We also compared to Bayesian neural network models which uses Stochastic Gradient MCMC \cite{Welling2011BayesianLV} and variational Bayes with horse shoe prior \cite{ghosh2018structured}. For all the neural network models, viz, artificial neural network (ANN), Stochastic Gradient MCMC Bayesian neural network (SG-MCMC) and variational Bayesian neural network with horse shoe prior (VBNN-HS), we fix the choice of the number of nodes at $k_n=10$ with a single hidden layer. 
For  variational Bayes neural network with horse shoe prior, we use the default  set up  as available in \cite{Ghosh_code}.

Table \ref{Table:different classifer} provides the training and testing accuracy and their empirical standard errors for all different methods under consideration. Learning from the previous section, the optimal hypter-parameters for BBVI and BBVI-CV are used as S=1000, $\rho_t=0.001$ and $S=200$, $\rho_t=0.001$, respectively.
In terms of the mean and standard deviation of testing accuracy, the best models are BBVI and BBVI-CV (almost equal performance). Apparently, the VBNN-HS suffers from over-fitting issue. LR, SVM, ANN and SG-MCMC have considerably larger standard errors for testing accuracy. Note, the BBVI and BBVI-CV algorithms have been fine tuned to perform optimally. One might observe an improvement in the performance of Stochastic Gradient MCMC Bayesian neural network and variational Bayesian neural network with horse shoe prior by optimally choosing their tuning parameters.  However studying that is beyond the scope of this paper as they are different methodology and the underlying statistical theories are not well established. 


\begin{table}[!h]
\small
\centering
\scalebox{1.35}{
\renewcommand{\arraystretch}{0.6}%
\begin{tabular}{| l  c  c |}
\hline
Classifier & Training accuracy (\%) & Testing accuracy (\%) \\\hline
LR & $82.1 \pm 2.5$ & $70.9 \pm 5.5 $\\\hline
 SVM & $80.32 \pm 2.2$ & $70.6 \pm 5.5$ \\\hline
ANN & $82.0 \pm 5.6$ & $74.1 \pm 6.8$ \\\hline
SG-MCMC & $80.8 \pm 4.6$ & $73.5 \pm 5.9$ \\\hline
BBVI& $79.46 \pm 1.7$ & $75.89 \pm 0.56$ \\\hline
BBVI-CV & $80.2 \pm 4.9$ & $75.71 \pm 0.61$ \\\hline
VBNN-HS  &  97.0$\pm 0.6$ & $71.6 \pm 4.16$\\\hline
\end{tabular}
}
 \caption{{ Performance for different classifiers. LR: Logistic regression. SVM: Support vector machine. ANN: Frequentist artificial neural network. SG-MCMC: Stochastic gradient MCMC Bayesian neural network. VBNN-HS: Variational Bayes neural network with horseshoe prior.}
}
 \label{Table:different classifer}
\end{table}

\section{Discussion and Conclusion} \label{Sec:Discussion}
The theoretical rigour and computational detail for variational Bayes neural network classifier presented in this article is  novel and unique contribution to statistical literature. Although the variational Bayes is popular in machine learning, neither the computational method nor the statistical properties are well understood for complex modeling such as neural networks. We characterize the prior distributions and the variational family for consistent Bayesian estimation. The theory also quantifies the loss due to VB numerical approximation compared to the true posterior distribution. For practical implementation, we reveal that the algorithm may not be as simple and straightforward as it sounds in computer science literature, rather it requires careful crafting on several parameters associated in various steps. Nevertheless, the computation could be quite faster compared to popular Monte Carlo Markov Chain procedure of approximating the posterior distributions. 

Although this article builds the framework on a  single layer neural networks model with simplistic prior structure, the detail statistical theory and computational methodology are quite involved. This investigation opens up possibility of exploring much wider class of models and priors. For example, shrinkage priors, such as double exponential and horseshoe priors can be explored for building sparse neural networks or one can experiment with various other variational families. However, their computational details and associated statistical properties are not immediate. We hope this research will accelerate further development of statistical and computational foundation for variational inference in general machine learning research.

\newpage

\section{Appendix}
\label{sec:append}

\section*{(A) Gradients of Variational family of BBVI}
In this section, we derive the analytical solution for BBVI with score function estimator.
 With $q$ as in \eqref{e:VIF}, for $j=1, \cdots, K(n)$
$$
\nabla_{m_{jn}}\mathcal{L}_{\mathcal{V}_q}=E_{q(.|\mathcal{V}_q)}\left[\left(\frac{\theta_{jn}-m_{jn}}{s_{jn}^2}\right)(\log \pi(\bTheta_n,\boldy_n,\boldX_n)-\log q(\bTheta_n|{\mathcal{V}_q}))\right]
$$
$$
g^{m_{jn}}=\frac{1}{S}\sum_{w=1}^S\left(\frac{\theta_{jn}[w]-m_{jn}}{s_{jn}^2}\right)\left(\log \pi(\btheta_n[w],\boldy_n,X_n)-\log q(\btheta_n[w]|{\mathcal{V}_q})\right)
$$

$$
\nabla_{s_{jn}}\mathcal{L}_{\mathcal{V}_q}=E_{q(.|\mathcal{V}_q)}\left[\left(\frac{(\theta_{jn}-m_{jn})^2}{s_{jn}^3}-\frac{1}{s_{jn}}\right)(\log \pi(\bTheta_n,\boldy_n,\boldX_n)-\log q(\bTheta_n|{\mathcal{V}_q}))\right]
$$
$$
g^{s_{jn}}=\frac{1}{S}\sum_{w=1}^S\left(\frac{(\theta_{jn}[w]-m_{jn})^2}{s_{jn}^3}-\frac{1}{s_{jn}}\right)\left(\log \pi(\btheta_n[w],\boldy_n,\boldX_n)-\log q(\btheta_n[w]|{\mathcal{V}_q})\right)
$$
With $s_{jn}=\log(1+e^{r_{jn}})$
$$
\nabla_{r_{jn}}\mathcal{L}_{\mathcal{V}_q}=\left(\frac{e^{r_{jn}}}{1+e^{r_{jn}}}\right)\left( \nabla_{s_{jn}}\mathcal{L}_{\mathcal{V}_q}|_{s_{jn}=\log(1+e^{r_{jn}})}\right)$$
$$
g^{r_{jn}}=\left(\frac{e^{r_{jn}}}{1+e^{r_{jn}}}\right) \left(g^{s_{jn}}|_{s_{jn}=\log(1+e^{r_{jn}})}\right)
$$

\section*{(B) Consistency of the variational posterior}

\subsection{Definitions}

\begin{definition}
\label{def:norms}
For a vector $\balpha$ and a function $g$,
\begin{enumerate}
    \item $||\balpha||_1=\sum_i |\alpha_i|$, $||\balpha||_2=\sqrt{\sum_i \alpha_i^2}$, $||\balpha||_\infty=\max_i |\alpha_i|$. 
    \item $||g||_1=\int_{\boldx \in \chi} |g(\boldx)|d\boldx $, $||g||_2=\sqrt{\int_{\boldx \in \chi} g(\boldx)^2d\boldx}$, $||g||_\infty=\sup_{\boldx \in \chi} |g(\boldx)|$
\end{enumerate}
\end{definition}

\begin{definition} \label{def:hell-brack}
For any two functions $l$ and $u$, define the
bracket $[l,u]$ as the set of all functions $f$ such that $l\leq  f\leq u$. Let $||.||$ be a metric. Define an $\varepsilon-$bracket as a bracket with $||u-l||\leq \varepsilon$. Define the bracketing number of a set of functions
$\mathcal{F}^*$ as the minimum number of $\varepsilon-$brackets needed to cover $\mathcal{F}^*$, and denote it by $N_{[]}(\varepsilon,\mathcal{F}^*,||.||)$. Finally, the
bracketing entropy, denoted $H_{[]}(\varepsilon,\mathcal{F}^*,||.||)$, is the natural logarithm of the bracketing number \cite{POL1990}.
\end{definition}

\subsection{Lemmas}

\begin{lemma}
\label{lem:h-bd}
Let $H_{[]}(u,\widetilde{\mathcal{F}}_n,||.||_2)\leq K(n)\log (M_n/u)$
then	$$\int_0^{\varepsilon}H_{[]}(u,\widetilde{\mathcal{F}}_n,||.||_2)du\lesssim \varepsilon\sqrt{K(n)(\log M_n-\log \varepsilon)}$$
where $H_{[]}(u,\widetilde{\mathcal{F}}_n,||.||_2)$ is defined in Definition \ref{def:hell-brack}.
\end{lemma}

\noindent \textit{Proof.} See proof of lemma 7.12 in \cite{BHAT2020}.

\begin{lemma}
	\label{lem:c-bound-0}
Suppose $q$ satisfies$$\int d_{\text{KL}}(\ell_0,\ell_{\btheta_n}) q(\btheta_n) d\btheta_n\leq \varepsilon,$$ then for any $\nu>0$,
	$$P_0^n\left(\left|\int q(\btheta_n) \log  \frac{L(\btheta_n)}{L_0}d\btheta_n\right|\geq n\nu\right)\leq \frac{\varepsilon}{\nu}$$
\end{lemma}

\noindent \textit{Proof.} See proof of lemma 7.11 in \cite{BHAT2020}.

\begin{lemma}	
	\label{lem:b-bound-0}
	Suppose $\mathcal{N}_\varepsilon=\{\btheta_n: d_{\text{KL}}(\ell_0,\ell_{\btheta_n})<\varepsilon\}$ and $p(\btheta_n)$ satisfies
	\begin{equation*}
	\label{e:nk-prior}
	\int_{\mathcal{N}_\varepsilon} p(\btheta_n)d\btheta_n\geq e^{-n\varepsilon}, n\to \infty
	\end{equation*}
	then for any $\nu>0$,
\begin{equation*}
\label{e:b-bound-0}
P_0^n\left(	\left|\log  \int \frac{L(\btheta_n)}{L_0}p(\btheta_n)d\btheta_n\right|\geq n \nu\right)\leq \frac{2\varepsilon}{\nu}
\end{equation*}
\end{lemma}

\noindent \textit{Proof.} See proof of lemma 7.10 in \cite{BHAT2020}.

\begin{lemma}
	\label{lem:pp-bound-0}
	Suppose,  $p(\btheta_n)$ satisfies 
	$$\int_{\mathcal{F}_n^c} p(\btheta_n) d\btheta_n \leq e^{-n\varepsilon }, n \to \infty$$
	for any $\varepsilon>0$. Then, 	for every $\tilde{\varepsilon}<\varepsilon$.
	$$P_0^n\left( \int_{\btheta_n \in \mathcal{F}_n^c}\frac{L(\btheta_n)}{L_0}p(\btheta_n)d\btheta_n\geq  e^{-n\tilde{\varepsilon} }\right) \leq e^{-n(\varepsilon-\tilde{\varepsilon})}$$
\end{lemma}

 \noindent \textit{Proof.} See proof of lemma 7.14 in \cite{BHAT2020}.

\begin{lemma}
	\label{lem:theta-bound}
	Let $\eta_{\boldt_n}(\boldx)=\beta^{\boldt}_{0}+\sum_{j=1}^{k_n}\beta^{\boldt}_{j}\psi({\gamma^{\boldt}_{j}}^\top \boldx)$ be a fixed neural network satisfying
$$|\theta_{jn}-t_{jn}|\leq \varepsilon,\:\: j=1, \cdots, K(n).$$
Then,	$$\int_{\boldx \in [0,1]^p} |\eta_{\btheta_n}(\boldx)- \eta_{\boldt_n}(\boldx)|dx\leq 8\left (K(n)+(p+1)\sum_{j=1}^{K(n)}|t_{in}|\right)\varepsilon$$
\end{lemma}

\noindent \textit{Proof.} See proof of lemma 7.2 in \cite{BHAT2020}.

\begin{lemma}
	\label{lem:h-bound}
If  $|\eta_0(\boldx)-\eta_{\btheta_n}(\boldx)|\leq \varepsilon$, then $|h_{\btheta_n}(\boldx)|\leq 2\varepsilon$  where $$h_{\btheta_n}(\boldx)= \sigma(\eta_0(\boldx))(\eta_0(\boldx)-\eta_{\btheta_n}(\boldx))+ \log(1-\sigma(\eta_0(\boldx))) -\log (1-\sigma(\eta_{\btheta_n}(\boldx)))$$
\end{lemma}

\noindent \textit{Proof:}
Note that,
\begin{align*}
|h_{\btheta_n}(\boldx)|&\leq |\sigma(\eta_0(\boldx))| |\eta_0(\boldx)-\eta_{\btheta_n}(\boldx)|+|\log (1-\sigma(\eta_0(\boldx))-\log(1-\sigma(\eta_{\btheta_n}(\boldx))|\\
&\leq |\eta_0(\boldx)-\eta_{\btheta_n}(\boldx)| +\left|\log\left(1+\sigma(\eta_0(\boldx))(e^{\eta_{\btheta_n}(\boldx)-\eta_0(\boldx)}-1)\right)\right|\\
&\leq 2|\eta_0(\boldx)-\eta_{\btheta_n}(\boldx)|
\end{align*}
where the second step follows by using $\sigma(x)=e^{x}/(1+e^x) \leq 1$ and  the proof of the third step is shown below.

\noindent Let $p=\sigma(\eta_0(\boldx))$, then $0\leq p \leq 1$ and $ r=\eta_{\btheta_n}(\boldx)-\eta_0(\boldx)$, then
\begin{align*}
\left|\log\left(1+\sigma(\eta_0(\boldx))(e^{\eta_{\btheta_n}(\boldx)-\eta_0(\boldx)}-1)\right)\right|=\left|\log\left(1+p(e^r-1)\right)\right|
\end{align*}
\begin{align*}
r>0&:\hspace{2mm}|\log (1+p(e^r-1))|=\log(1+p(e^r-1))\leq \log(1+(e^r-1))=r=|r|\\
r<0&:\hspace{2mm} |\log (1+p(e^r-1))|=-\log(1+p(e^r-1))\leq -\log(1+(e^r-1))=-r=|r|
\end{align*} 
 
 \begin{lemma}
 	\label{lem:d2h-bound}
 	For  $$h(\btheta_n)= \int_{\boldx \in [0,1]^p} |\eta_{\btheta_n}\ \left( \sigma(\eta_0(\boldx))(\eta_0(\boldx)-\eta_{\btheta_n}(\boldx))+ \log(1-\sigma(\eta_0(\boldx))) -\log (1-\sigma(\eta_{\btheta_n}(\boldx)))\right)d \boldx$$
 $$\sum_{j=1}^{K(n)}| (\nabla^2 h(\btheta_n))_{jj}|\leq  (2(p+1)+1)(K(n)+1)+2(p+1)
 ||\btheta_n||_2^2$$
 where $A_{jj}$ denotes the $j^{\text{th}}$ diagonal entry of a matrix.
 \end{lemma}
 
\noindent \textit{Proof.} First note that \begin{align*}
 \nabla^2 h(\btheta_n)&=-\int_{\boldx \in [0,1]^p}  \underbrace{(\sigma(\eta_0(\boldx))+\sigma(\eta_{\btheta_n}(\boldx)))}_{g_1(\boldx)}\nabla^2 \eta_{\btheta_n}(\boldx) d \boldx\\
 &-\int_{\boldx \in [0,1]^p}  \underbrace{ (\sigma(\eta_{\btheta_n}(\boldx)))(1-\sigma(\eta_{\btheta_n}(\boldx)))}_{g_2(\boldx)}\nabla \eta_{\btheta_n}(\boldx) \nabla \eta_{\btheta_n}(\boldx)^{\top} d\boldx
 \end{align*}
 For $r=\lfloor(j-k_n-1)/p\rfloor , r'= (j-k_n-1)\text{ mod }p$
 $$-b_{jj}=\begin{cases}
 \int g_2(\boldx) d\boldx, & r=0, r'=0\\
 \int g_2(\boldx) (\psi(\gamma_{r'}^\top \boldx))^2 d\boldx, &r=0, r'=1, \cdots, k_n\\
 \int g_1(\boldx)\beta_r\psi''(\gamma_{r}^\top \boldx) d\boldx+\int  g_2(\boldx) \beta_r^2  (\psi'(\gamma_{r}^\top \boldx))^2  d\boldx, & r=1, \cdots, k_n, r'=0\\
 \int g_1(\boldx)\beta_r\psi''(\gamma_{r}^\top \boldx)x_{r'}^2 d\boldx+\int  g_2(\boldx) \beta_r^2  (\psi'(\gamma_{r}^\top \boldx))^2 x_{r'}^2 d\boldx, & r=1, \cdots, k_n, r'=1, \cdots, p\\
 \end{cases}$$
 where the integral is over the set $\boldx \in [0,1]^p$.   
 
\noindent  Note, $\psi(u) \leq 1$, $|g_1(\boldx)|\leq 2$ and $|g_2(\boldx)| \leq 1$. Also, $\psi(u), \psi'(u), \psi''(u), |x_{r'}^2| \leq 1$ which implies 
 \begin{align*}
 \sum_{j=1}^{K(n)}|b_{jj}|&\leq (k_n+1)+(p+1)\sum_{j=1}^{k_n+1}(2|\beta_j|+|\beta_j|^2)\\
 &\leq (k_n+1)+(p+1)\sum_{j=1}^{k_n+1}(2+2|\beta_j|^2)\\
&\leq (2(p+1)+1)(K(n)+1)+2(p+1)||\btheta_n||_2^2
 \end{align*}
where the second inequality in the above step uses $|x|<x^2+1$. 
 \begin{lemma}
 	\label{lem:e-bound-0}
 	Let, $\widetilde{\mathcal{F}}_n=\{\sqrt{\ell}: \ell_{\btheta_n}(y,\boldx), \btheta_n \in \mathcal{F}_n\}$ where $\ell_{\btheta_n}(y,\boldx)$ is same as in \eqref{e:lik-one} and $\mathcal{F}_n$ is same as in \eqref{e:Fn-def}.
 	Then,
 	$$\int_{\varepsilon^2/8}^{\sqrt{2}\varepsilon}\sqrt{H_{[]}(u,\widetilde{\mathcal{F}}_n,||.||_2)}du \lesssim  \varepsilon \sqrt{2K_n (\log K_n+2\log C_n-\log \varepsilon)} $$
where $H_{[]}(u,\widetilde{\mathcal{F}}_n,||.||_2)$ is defined in Definition \ref{def:hell-brack}.
 \end{lemma}

\noindent \textit{Proof.} In this proof, let $\btheta=\btheta_n$.	Note, by lemma 4.1 in \cite{POL1990},
	$$N(\varepsilon,\mathcal{F}_n,||.||_\infty)\leq \left(\frac{3C_n}{\varepsilon}\right)^{K(n)}.$$
 	\noindent For $\btheta_1, \btheta_2 \in \mathcal{F}_n$, let $\widetilde{\ell}(u)=\sqrt{\ell_{u\btheta_1+(1-u)\btheta_2}(\boldx,y)}$.

 	Following equation (52) in \cite{BHAT2020}, we get
 	\begin{align}
 	\label{e:l-bd-0}
 \sqrt{\ell_{\btheta_1}(\boldx,y)}-\sqrt{\ell_{\btheta_2}(\boldx,y)}&\leq K(n)\sup_{j}  \Big|\pp{\widetilde{\ell}}{\theta_j}\Big|||\btheta_1-\btheta_2||_{\infty}  	\leq F(\boldx,y)||\btheta_1-\btheta_2||_{\infty}
 	\end{align}
 	where the upper bound $F(\boldx,y)=K(n)C_n/2$. This is because $|\partial \widetilde{\ell}/\partial \theta_j|$, the derivative of $\sqrt{\ell}$ w.r.t.  is bounded above by $|\partial \eta_{\btheta}(\boldx)/\partial \theta_j|$ as shown below.
 	\begin{align*}
 \left|	\pp{\widetilde{\ell}}{\theta_j}\right|&=\left|\frac{1}{2}\pp{\eta_{\btheta}(\boldx)}{\theta_j}\left(y-\frac{e^{\eta_{\btheta}(\boldx)}}{1+e^{\eta_{\btheta}(\boldx)}}\right)\sqrt{e^{(y\eta_{\btheta}(\boldx)-\log(1+e^{\eta_{\btheta}(\boldx)}))}}\right|\\
 &\leq \frac{1}{2}	\pp{\eta_{\btheta}(\boldx)}{\theta_j}\left(\frac{e^{\eta_{\btheta}(\boldx)}}{1+e^{\eta_{\btheta}(\boldx)}}\right)^{1/2}\left(\frac{1}{1+e^{\eta_{\btheta}(\boldx)}}\right)^{1/2}
 	\end{align*} 
 Thus, using $e^{\eta_{\btheta}(\boldx)}/(1+e^{\eta_{\btheta}(\boldx)})$, $1/(1+e^{\eta_{\btheta}(\boldx)})\leq 1$, we get
 $$2	\left|\pp{\widetilde{\ell}}{\theta_j}\right|\leq \left|	\pp{\eta_{\btheta}(\boldx)}{\theta_j}\right|\leq \begin{cases}
  1, & \theta_j=\beta_r \text{ for some $r=0, \cdots, k_n$}\\
 |\beta_r \psi'(\gamma_r^\top x) x_{r'}|, & \theta_j=\gamma_{rr'} \text{ for some $r=0, \cdots, k_n, r'=0, \cdots, p$}
 \end{cases}$$
 Using $|\psi'(u)|\leq 1$, $|x_{r'}|\leq 1$, $|\beta_r|\leq C_n$, the bound $F(\boldx,y)$ follows.
 
 \noindent In view of \eqref{e:l-bd-0}  and theorem 2.7.11 in \cite{VW1996}, we have
 	$$N_{[]}(\varepsilon, \widetilde{\mathcal{F}}_n, ||.||_2)\leq \left(\frac{3K(n)C_n^2}{2\varepsilon}\right)^{K(n)}\implies H_{[]}(\varepsilon, \widetilde{\mathcal{F}}_n, ||.||_2)\lesssim K(n) \log \frac{K(n)C_n^2}{\varepsilon}$$
 	where $N_{[]}$ and $H_{[]}$ denote the bracketing number and bracketing entropy as in definition \ref{def:hell-brack}.
 	
 	Using, lemma \ref{lem:h-bd} with $M_n=K(n)C_n^2$, we get
 	$$\int_0^{\varepsilon} \sqrt {H_{[]}(u, \widetilde{\mathcal{F}}_n, ||.||_2)} du\lesssim \varepsilon \sqrt{K(n)(\log K(n)C_n^2-\log \varepsilon)}$$
 	Therefore, 
 	\begin{align*}
 	    \int_{\varepsilon^2/8}^{\sqrt{2}\varepsilon} H_{[]}(u, \widetilde{\mathcal{F}}_n, ||.||_2) du&\leq\int_{0}^{\sqrt{2}\varepsilon} {H_{[]}(u, \widetilde{\mathcal{F}}_n, ||.||_2)} du\\
 	    &\lesssim \sqrt{2}\varepsilon \sqrt{K(n)(\log K(n)C_n^2-\log \sqrt{2} \varepsilon)} 
 	\end{align*}
 	The proof follows by noting $\log \sqrt{2}\varepsilon \geq \log \varepsilon$.

\subsection{Propositions}

\begin{proposition}
	\label{e:kl-q-p}
Let $q(\btheta_n)=MVN(\boldt_n,I_{K(n)}/\sqrt{n})$ and  $p(\btheta_n)=MVN(\bmu_n,\bSigma_n)$, where $\bSigma_n=\text{diag}(\bzeta_n)$ and $\bzeta^*_n=1/\bzeta_n$.
Let $n\epsilon_n^2 \to \infty$,   $K(n)\log n =o(n\epsilon_n^2)$, $||\boldt_n||_2^2=o(n\epsilon_n^2)$, $||\bmu_n||_2^2=o(n\epsilon_n^2)$,  then for any $\nu>0$,   
     $$d_{\rm KL}(q,p)\leq n\epsilon_n^2\nu$$
     provided $||\bzeta_n||_\infty=O(n)$, $||\bzeta^*_n||_\infty=O(1)$. 
\end{proposition}

\noindent \textit{Proof:}
\begin{align} 
\nonumber d_{\rm KL}(q,p)&=\sum_{j=1}^{K(n)}\left(\log \sqrt{n}\zeta_{jn}+\frac{1}{n\zeta_{jn}^2}+\frac{(t_{jn}-\mu_{jn})^2}{\zeta_{jn}^2}-\frac{1}{2}\right)\\
\nonumber& \leq \frac{K(n)}{2}(\log n-1)+\sum_{j=1}^{K(n)}\log \zeta_{jn}+\frac{1}{n}\sum_{j=1}^{K(n)}\frac{1}{\zeta_{jn}^2}+2\sum_{j=1}^{K(n)}\frac{t_{jn}^2}{\zeta_{jn}^2}+2\sum_{j=1}^{K(n)}\frac{\mu_{jn}^2}{\zeta_{jn}^2}-\frac{K(n)}{2}\\
\nonumber&\leq \frac{K(n)}{2}(\log n-1)+K(n)\log ||\bzeta_n||_\infty+\frac{K(n)}{n}||\bzeta_n^*||_\infty\\
&+2||\boldt_n||_2^2||\bzeta_n^*||_\infty+2||\bmu_n||_2^2||\bzeta_n^*||_\infty=o(n\epsilon_n^2)
\end{align}
where the second last inequality uses $\bzeta^*_n=1/\bzeta_n$.
The last equality follows since $||\bzeta_n||=O(n)$, $||\bzeta_n^*||_\infty=O(1)$, $K(n)\log n=o(n\epsilon_n^2)$, $||\boldt_n||_2^2=o(n\epsilon_n^2)$ and $||\bmu_n||_2^2=o(n\epsilon_n^2)$.

\begin{proposition}
	\label{lem:kl-bound-v}
Let $p(\btheta_n)$ as in \eqref{e:prior}.
Define 
\begin{eqnarray}
\label{e:Nk-def-v}
 \mathcal{N}_\varepsilon=\left\{\btheta_n:d_{\rm KL}(\ell_0,\ell_{\btheta_n})<\varepsilon\right\}
\end{eqnarray}
where
$$d_{\rm KL}(\ell_0,\ell_{\btheta_n})=\int_{\boldx \in [0,1]^p} \left( \sigma(\eta_0(\boldx))(\eta_0(\boldx)-\eta_{\btheta_n}(\boldx))+ \log\frac{1-\sigma(\eta_0(\boldx))}{1-\sigma(\eta_{\btheta_n}(\boldx))}\right) d\boldx$$
Let $||\eta_0-\eta_{\boldt_n}||_\infty\leq \varepsilon\epsilon_n^2/4$, $n\epsilon_n^2 \to \infty$. If $K(n)\log n =o(n\epsilon_n^2)$, $||\boldt_n||_2^2=o(n\epsilon_n^2)$, $||\bmu_n||_2^2=o(n\epsilon_n^2)$,  then  $$ \int_{\btheta_n\in N_{\varepsilon \epsilon_n^2}} p(\btheta_n)d\btheta_n \geq e^{-n\epsilon_n^2\nu} \hspace{3mm}\forall \:\: \nu>0$$
        provided $||\bzeta_n||_\infty=O(n)$, $||\bzeta^*_n||_\infty=O(1)$ where $\bzeta^*_n=1/\bzeta_n$.
 \end{proposition}

\noindent \textit{Proof.}
\noindent Let $\eta_{\boldt_n}(\boldx)=\beta^{\boldt}_{0}+\sum_{j=1}^{k_n}\beta^{\boldt}_{j}\sigma({\gamma^{\boldt}_{j}}^\top \boldx)$ be the neural network such that
	\begin{equation}
	\label{e:n-delta-1-v}
	||\eta_{\boldt_n}-\eta_0||_1\leq \frac{\varepsilon \epsilon_n^2}{4}
	\end{equation}

\noindent Such a neural network exists since $||\eta_0-\eta_{\boldt_n}||_1\leq ||\eta_0-\eta_{\boldt_n}||_\infty\leq \varepsilon \epsilon_n^2/4$.

Next define neighborhood $\mathcal{M}_{\varepsilon}$ as follows
	\begin{align*}
	\mathcal{M}_{\varepsilon \epsilon_n^2}=\left\{\btheta_n:|{\theta}_{jn}-{t}_{jn}|<\frac{\varepsilon\epsilon_n^2}{8(K(n)+(p+1)||\boldt_{n}||_1)}, j=1,\cdots, K(n)\right\}
	\end{align*}
For every $\btheta_n \in \mathcal{M}_{\varepsilon \epsilon_n^2}$, by lemma \ref{lem:theta-bound}, we have
\begin{align}
\label{e:n-delta-2-v}
||\eta_{\btheta_n}-\eta_{\boldt_n}||_1 \leq \frac{\varepsilon\epsilon_n^2}{2}
\end{align} 
Combining \eqref{e:n-delta-1-v} and \eqref{e:n-delta-2-v}, we get for $\btheta_n\in \mathcal{M}_{\varepsilon \epsilon_n^2}$, $||\eta_{\btheta_n}-\eta_{0}||_1 \leq \varepsilon\epsilon_n^2/2$.

\noindent This, in view of lemma \ref{lem:h-bound},
$d_{\rm KL}(\ell_0,\ell_{\btheta_n}) \leq \varepsilon \epsilon_n^2$.
	
\noindent Let  $\btheta_n \in \mathcal{N}_{\varepsilon \epsilon_n^2}$ for every $\btheta_n \in \mathcal{M}_{\varepsilon \epsilon_n^2}$. Therefore, 
	$$\int_{\btheta_n \in \mathcal{N}{\varepsilon \epsilon_n^2}} p(\btheta_n)d\btheta_n\geq \int_{\btheta_n \in \mathcal{M}_{\varepsilon \epsilon_n^2}} p(\btheta_n)d\btheta_n$$
Let $\delta_n=\varepsilon\epsilon_n^2/(8(K(n)+(p+1)||\boldt_{n}||_1))$, then
\begin{align}
\label{e:the-bound}
\nonumber\int_{\btheta_n \in \mathcal{M}_{\varepsilon \epsilon_n^2}}p(\btheta_n)d\btheta_n&=\prod_{j=1}^{K(n)}\int_{t_{jn}-\delta_{n}}^{t_{jn}+\delta_{n}}\frac{1}{\sqrt{2\pi\zeta_{jn}^2}}e^{-\frac{(\theta_{jn}-\mu_{jn})^2}{2\zeta_{jn}^2}}d\theta_{jn}\\
\nonumber &= \prod_{j=1}^{K(n)}\frac{2\delta_{n}}{\sqrt{2\pi\zeta_{jn}^2}}e^{-\frac{(\tilde{t}_{jn}-\mu_{jn})^2}{2\zeta_{jn}^2}},\:\: \tilde{t}_{jn}\in [t_{jn}-\delta_{n},t_{jn}+\delta_{n}]\\
 &=\prod_{j=1}^{K(n)}e^{-\left(-\frac{1}{2}\log \frac{2}{\pi}-\log \delta_n+\log \zeta_{jn}+\frac{(\tilde{t}_{jn}-\mu_{jn})^2}{2\zeta_{jn}^2}\right)} 
 \end{align}
where the second last equality holds by mean value theorem.

\noindent Note that $\tilde{t}_{jn} \in [t_{jn}-1,t_{jn}+1]$ since $\delta_n \to 0$, therefore
\begin{align*}
\frac{(\tilde{t}_{jn}-\mu_{jn})^2}{2\zeta_{jn}^2} \leq \frac{\max( (t_{jn}-\mu_{jn}-1)^2,(t_{jn}-\mu_{jn}+1)^2)}{2\zeta_{jn}^2}\leq \frac{(t_{jn}-\mu_{jn})^2}{\zeta_{jn}^2}+\frac{1 }{\zeta_{jn}^2}
\end{align*}
where the last inequality follows since $(a+b)^2\leq 2(a^2+b^2)$. Therefore
\begin{align}
\label{e:p1-ub}
\nonumber \sum_{j=1}^{K(n)}\frac{(\tilde{t}_{jn}-\mu_{jn})^2}{2\zeta_{jn}^2}&\leq  2\sum_{j=1}^{K(n)}\frac{t_{jn}^2}{\zeta_{jn}^2}+ 2\sum_{j=1}^{K(n)}\frac{\mu_{jn}^2}{\zeta_{jn}^2}+\sum_{j=1}^{K(n)}\frac{1}{\zeta_{jn}^2}\\
&\leq 2 (||\boldt_n||_2^2+||\bmu_n||_2^2+1) ||\bzeta_n^*||_\infty  \leq n\nu\epsilon_n^2
\end{align}
since $||\boldt_n||_2^2=o(n\epsilon_n^2)$, $||\bmu_n||_2^2=o(n\epsilon_n^2)$ and $||\bzeta_n^*||_\infty=O(1)$ and $n\epsilon_n^2 \to \infty$.
Also, 
\begin{align*}
-\log \delta_n +\log \zeta_{jn}&=\log 8 +\log (K(n)+(p+1)||\boldt_n||_1)+\log \zeta_{jn}-\log \varepsilon\epsilon_n^2\\
&\leq \log 8 +\log (K(n)+(p+1)\sqrt{K(n)}||\boldt_n||_2)+\log \zeta_{jn}-\log \varepsilon-2\log  \epsilon_n\\
&\leq \log 8 +\log K(n)+\log\left(1+||\boldt_n||_2\right)+\log \zeta_{jn}-\log \varepsilon-2\log  \epsilon_n
\end{align*}
where the second inequality is an outcome of Cauchy Schwartz and the third inequality follows since $(p+1)\leq \sqrt{K(n)}$, $n\to \infty$. Therefore,
\begin{align}
\label{e:p2-ub}
\nonumber\sum_{j=1}^{K(n)}-\frac{1}{2}\log \frac{2}{\pi}-\log \delta_n +\log \zeta_{jn}&\leq K(n)\log 8+K(n)\log K(n)++K(n)\log  (1+||\boldt_n||_2)\\
&+K(n)\log ||\bzeta_n||_\infty-K(n)\log \varepsilon-2K(n)\log \epsilon_n\leq n\nu\epsilon_n^2
\end{align}
where the last inequality follows since $K(n)\log n=o(n\epsilon_n^2)$, $||\zeta_n||_\infty=O(n)$, $||\boldt_n||_2=o(\sqrt{n}\epsilon_n)=o(n)$ and $1/n\epsilon_n^2=o(1)$ which implies $-2\log \epsilon_n=o(\log n)$. 

\noindent Combining \eqref{e:p1-ub} and \eqref{e:p2-ub} and replacing \eqref{e:the-bound}, the proof follows.

\begin{proposition}
	\label{lem:f-bound}
Let $q(\btheta_n) \sim MVN(\boldt_n,I_{K(n)}/\sqrt{n})$.  Define
	$$h(\btheta_n)= \int_{\boldx \in [0,1]^p} \left( \sigma(\eta_0(\boldx))(\eta_0(\boldx)-\eta_{\btheta_n}(\boldx))+ \log \frac{1-\sigma(\eta_0(\boldx))}{1-\sigma(\eta_{\btheta_n}(\boldx))}\right)d\boldx$$
Let $||\eta_0-\eta_{\boldt_n}||_\infty\leq \varepsilon\epsilon_n^2/4$ where $n\epsilon_n^2 \to \infty$.
If $K(n)\log n=o(n\epsilon_n^2)$, $||\boldt_n||_2^2=o(n\epsilon_n^2)$,  then   $$\int h(\btheta_n)q(\btheta_n)d\btheta_n \leq  \varepsilon\epsilon_n^2$$
provided $||\bzeta_n||_\infty=O(n)$, $||\bzeta^*_n||_\infty=O(1)$ where $\bzeta^*_n=1/\bzeta_n$.

\end{proposition}

\noindent \textit{Proof.} Since $h(\btheta_n)$ is a KL-distance, $h(\btheta_n)>0$. We shall thus establish an upper bound.

\noindent Let $A=\{\btheta_n:\cap_{j=1}^{K(n)}|\theta_{jn}-t_{jn}|\leq \sqrt{\varepsilon \epsilon_n^2/K(n)} \}$, then
\begin{equation}
\label{e:h-split-1}
    \int h(\btheta_n)q(\btheta_n)d\btheta_n=\int_{A}h(\btheta_n)q(\btheta_n)d\btheta_n+\int_{A^c}h(\btheta_n)q(\btheta_n)d\btheta_n
\end{equation}
By Taylor expansion, the first term is equal to

\begin{align}
\label{e:h-bound-1}
\nonumber&=\int_A\left(h(\boldt_n)+(\btheta_n-\boldt_n)^{\top}\nabla h(\boldt_n)+\frac{1}{2}(\btheta_n-\boldt_n)^{\top}\nabla^2 h(\boldt_n)(\btheta_n-\boldt_n)\right)q(\btheta_n)d\btheta_n+o(\varepsilon \epsilon_n^2)\\
\nonumber&\leq |h(\boldt_n)|+\frac{1}{2}\int (\btheta_n-\boldt_n)^{\top}\nabla^2 h(\boldt_n)(\btheta_n-\boldt_n)q(\btheta_n)d\btheta_n+o(\varepsilon\epsilon_n^2)\\
\nonumber&=\frac{\varepsilon \epsilon_n^2}{2}+\frac{1}{2}\int (\btheta_n-\boldt_n)^{\top}\nabla^2 h(\boldt_n)(\btheta_n-\boldt_n)q(\btheta_n)d\btheta_n+o(\varepsilon \epsilon_n^2)\\
&=\frac{\varepsilon \epsilon_n^2}{2}+o(\varepsilon \epsilon_n^2) \leq \frac{3\varepsilon \epsilon_n^2}{4} 
\end{align}
where the second step holds because $q(\btheta_n)$ is symmetric around $\boldt_n$. The third step holds in view of  lemma \ref{lem:h-bound} and the fact that
$\boldt_n$ satisfies $||\eta_{\boldt_n}-\eta_0||_\infty \leq \varepsilon\epsilon_n^2/4$. 

The final step is justified next.
With $J=\{1,\cdots, K(n)\}$, let  $\nabla^2 h(\boldt_n)=((b_{jj'}))_{j\in J, j'\in J}$
\begin{align*}
	\int_A (\btheta_n-\boldt_n)^{\top}\nabla^2 h(\boldt_n)(\btheta_n-\boldt_n)q(\btheta_n)d\btheta_n&=\sum_{j=1}^{K(n)} b_{jj}\int_{|{\theta}_{jn}-t_{jn}|\leq \sqrt{\varepsilon \epsilon_n^2/K(n)}}(\theta_{jn}-t_{jn})^2 q(\theta_{jn})d\theta_{jn}
	\end{align*}
where the cross covariance terms disappear since $\theta_{jn}$'s are independent and $q(\btheta_n)$ is symmetric around $\boldt_n$.  Thus, 
	\begin{align*}
	\int_A (\btheta_n-\boldt_n)^{\top}\nabla^2 h(\boldt_n)(\btheta_n-\boldt_n)q(\btheta_n)d\btheta_n&\leq \sum_{j=1}^{K(n)}|b_{jj}|\int(\theta_{jn}-t_{jn})^2 q(\theta_{jn})d\theta_{jn}=\frac{1}{n}\sum_{j=1}^{K(n)}|b_{jj}| 
	\end{align*}
Using lemma \ref{lem:d2h-bound}, we get	
\begin{align}
\label{e:d2h-bound}
\nonumber	\int_A (\btheta_n-\boldt_n)^{\top}\nabla^2 h(\boldt_n)(\btheta_n-\boldt_n)q(\btheta_n)d\btheta_n&\leq \frac{1}{n}((2(p+1)+1)(K(n)+1)+2(p+1)||\boldt_n||^2_2)\\
	&=o(\varepsilon \epsilon_n^2)
\end{align}
where the last equality holds since $K(n)\log n=o(n\epsilon_n^2)$ and $||\boldt_n||_2^2=o(\varepsilon \epsilon_n^2)$.
We next handle the second term in \eqref{e:h-split-1}. Using  lemma \ref{lem:h-bound}, note that
\begin{align*}
	\int_{A^c} h(\btheta_n)q(\btheta_n)d\btheta_n&\leq 2\int_{A^c}  \left(\int_{\boldx \in [0,1]^p}  |\eta_0(\boldx)-\eta_{\btheta_n}(\boldx)|d\boldx\right) q(\btheta_n)d\btheta_n\\
	&\leq 2 \int_{\boldx \in [0,1]^p} |\eta_0(\boldx)|d\boldx \int_{A^c}q(\btheta_n)d\btheta_n+2 \int_{A^c}\int_{\boldx \in [0,1]^p} |\eta_{\btheta_n}(\boldx)|  d\boldx  q(\btheta_n) d\btheta_n 
\end{align*}
	First, note using $|\psi(u)|\leq 1$, we get $|\eta_{\btheta_n}(\boldx)|\leq \sum_{j=0}^{k_n}|\beta^{\boldt}_j|$. Thus, $|\eta_{\btheta_n}(\boldx)|\leq \sum_{j=0}^{k_n} |\beta^{\boldt}_{j}|+\sum_{j=0}^{k_n}|\beta_{j}-\beta^{\boldt}_j|$ which implies 
	
\begin{align}
\label{e:q-val}
\frac{1}{2}\int_{A^c} h(\btheta_n)q(\btheta_n)d\btheta_n
&\leq  Q(A^c)\left(\int_{\boldx \in [0,1]^p} |\eta_0(\boldx)| d\boldx+ \sum_{j=0}^{k_n} |\beta^{\boldt}_{j}|\right)+\int_{A^c}\left(\sum_{j=0}^{k_n} |\beta_{j}-\beta^{\boldt}_j|\right) q(\btheta_n)d\btheta_n
\end{align}
\noindent First note that $A^c=\cup_{j=1}^{K(n)}A_j^c$ where  $A_j=\{|\theta_{jn}-t_{jn}|\leq \sqrt{\varepsilon\epsilon_n^2/K_n}\}$. Therefore,
\begin{align}
\label{e:a-form}
\nonumber Q(A^c)&=Q(\cup_{j=1}^{K(n)}A_j^c)\leq\sum_{j=1}^{K(n)}Q(A_j^c)=\sum_{j=1}^{K(n)}\int_{|\theta_{jn}-t_{jn}|>\sqrt{\varepsilon \epsilon_n^2/K(n)}}q(\theta_{jn})d\theta_{jn}\\
&=2K(n)\left(1-\Phi\left(\sqrt{\frac{n \varepsilon \epsilon_n^2}{K(n)}}\right)\right)
\end{align}

\noindent Using \eqref{e:a-form} in the first term of \eqref{e:q-val}, we get
\begin{align}
\label{e:q-ub}
\nonumber Q(A^c)\left(\int_{\boldx \in [0,1]^p} |\eta_0(\boldx)| d\boldx+ \sum_{j=0}^{k_n} |\beta^{\boldt}_{j}|\right)&\lesssim  2(||\eta_0||_1 +||\boldt_n||_1)K(n)\left(1-\Phi\left(\sqrt{\frac{n \varepsilon \epsilon_n^2}{K(n)}}\right)\right)\\
\nonumber&\hspace{-10mm}\stackrel{\text{Cauchy Schwartz}}{\leq}  2(||\eta_0||_1 +\sqrt{K(n)}||\boldt_n||_2)K(n)\left(1-\Phi\left(\sqrt{\frac{n \varepsilon \epsilon_n^2}{K(n)}}\right)\right)\\
\nonumber&\leq  4 n\epsilon_n^2 K(n)\left(1-\Phi\left(\sqrt{\frac{n \varepsilon \epsilon_n^2}{K(n)}}\right)\right)\\
\nonumber&\leq  4 n K(n)\left(1-\Phi\left(\sqrt{\frac{n \varepsilon \epsilon_n^2}{K(n)}}\right)\right) \\
\nonumber&\hspace{-5mm}\stackrel{\text{Mill's ratio}}{\sim}  4 nK(n)\sqrt{\frac{K(n)}{n\varepsilon \epsilon_n^2}} e^{-\frac{n\varepsilon\epsilon_n^2}{2 K(n)}} \\
&\leq 4 nK(n)e^{-\frac{n \varepsilon\epsilon_n^2}{2 K(n)}}= o(\varepsilon\epsilon_n^2)
\end{align}
where the third step holds holds because $||\boldt_n||_2=o(\sqrt{n\epsilon_n^2})$ and  $\sqrt{K(n)}=o(\sqrt{n\epsilon_n^2})$ and $||\eta_0||_1$ is fixed and fourth step holds because $\epsilon_n^2\leq 1$. The last equality in the above step holds because
$$-\frac{n\epsilon_n^2}{K(n)}+\log K(n)+\log n -\log \varepsilon\leq  -\frac{n\epsilon_n^2}{K(n)}+3\log n =-\log n\left(\frac{n\epsilon_n^2}{K(n)\log n}-3\right) \to -\infty $$
where the first inequality holds since  $K(n)\leq n$.

\noindent For the second term in \eqref{e:q-val}, let $A_{\beta_j}=\{|\beta_{j}-\beta^{\boldt}_{j}|>\sqrt{\varepsilon \epsilon_n^2/K_n}\}$

\begin{align}
\label{e:beta-def}
 \nonumber   \int_{A^c}\left(\sum_{j=0}^{k_n}|\beta_j-\beta^{\boldt}_{j}|\right)q(\btheta_n)d\btheta_n&=\sum_{j=0}^{k_n}\int_{A^c}|\beta_j-\beta^{\boldt}_{j}|q(\btheta_n)d\btheta_n\\
 \nonumber  &\hspace{-10mm}=\sum_{j=0}^{k_n}\left(\int_{A^c\cap A_{\theta_j}} |\beta_j-\beta^{\boldt}_{j}|q(\btheta_n)d\btheta_n+\int_{A^c \cap A_{\beta_j}^c}|\beta_j-\beta^{\boldt}_{j}|q(\btheta_n)d\btheta_n\right)\\
 & \hspace{-10mm}\leq \sum_{j=0}^{k_n}\left( \int_{A_{\beta_j}} |\beta_j-\beta^{\boldt}_{j}|q(\beta_j)d\beta_j +E_{q(\beta_j)}|\beta_j-\beta^{\boldt}_j|\int_{\tilde{A}^c} q(\tilde{\btheta}_n)d\tilde{\btheta}_n\right) 
 \end{align} 
where $\tilde{\btheta}_n$ has all coordinates of  $\btheta_n$ except $\beta_j$ and $\tilde{A}^c$ is the union of all $A_j^c$ except $A_{\beta_j}^c$.
\begin{align}
\label{e:beta-def-1}
\sum_{j=0}^{k_n}    \int_{A_{\beta_j}} |\theta_j-\beta^{\boldt}_{j}|q(\beta_j)d\beta_j
\nonumber &=\sum_{j=0}^{k_n}\int_{|\beta_j-\beta^{\boldt}_{j}|>\sqrt{\varepsilon\epsilon_n^2/K(n)}}\sqrt{\frac{n}{2\pi}}|\beta_j-\beta^{\boldt}_{j}|e^{-\frac{n}{2}|\beta_j-\beta^{\boldt}_j|^2}d\beta_j\\
\nonumber&=\frac{2}{\sqrt{n}}\sum_{j=0}^{k_n}\int_{\sqrt{n\varepsilon\epsilon_n^2/K(n)}}^{\infty} \frac{u}{\sqrt{2\pi}}e^{-\frac{1}{2}u^2}du \\
&\leq 2K(n)e^{-\frac{n\varepsilon \epsilon_n^2}{2K(n)}}=o(\varepsilon \epsilon_n^2)
\end{align}
where the last equality is a consequence of \eqref{e:q-ub}.

Note $E_{q(\beta_j)}|\beta_j-\beta_j^{\boldt}|=\sqrt{2/\pi}(1/n)$. Thus
\begin{align}
\label{e:beta-def-2}
\nonumber\sum_{j=0}^{k(n)}E_{q(\beta_j)}|\beta_j-\beta^{\boldt}_j|\int_{\tilde{A}^c} q(\tilde{\btheta}_n)d\tilde{\btheta}_n& = \frac{2K(n)}{n\sqrt{2\pi}} Q(\tilde{A}^c)\\
\nonumber&\sim\frac{2K(n)^2}{n\sqrt{2\pi}} \left(1-\Phi\left(\sqrt{\frac{n\varepsilon}{K(n)}}\right)\right)\\
&\leq 2 K(n) \left(1-\Phi\left(\sqrt{\frac{n\varepsilon}{K(n)}}\right)\right)=o(\varepsilon \epsilon_n^2)
\end{align}
where the asymptotic equality  in the second line follows from \eqref{e:a-form}  because $Q(\tilde{A}^c)$ shares the same form as $Q(A^c)$. 
The third line follow from $K(n)\leq n$ and Relation \eqref{e:q-ub}.

Combining \eqref{e:q-ub}, \eqref{e:beta-def-1}, \eqref{e:beta-def-2}, we get \begin{equation}
\label{e:h-bound-2}
    \int_{A^c} h(\btheta_n)q(\btheta_n)d\btheta_n=o(\varepsilon\epsilon_n^2)\leq \frac{\varepsilon\epsilon_n^2}{4}
\end{equation}
This together with \eqref{e:h-bound-1} completes the proof.
\begin{proposition}
	\label{lem:p-bound-0}
Let $n\epsilon_n^2\to \infty$. Suppose   $p(\btheta_n)$ satisfies \eqref{e:prior} with $||\bmu_n||_2^2=o(n\epsilon_n^2)$ and  $||\bzeta_n||_\infty=O(n)$. Suppose for some $0<b<1$, $K(n)\log n=o(n^b\epsilon_n^2)$, then for  $C_n=e^{n^b \epsilon_n^2/K(n)}$ and $\mathcal{F}_n$ as in \eqref{e:Fn-def}, we have for any $\varepsilon>0$,		$$\int_{\btheta_n \in \mathcal{F}_n^c}p(\btheta_n)d\btheta_n\leq e^{- n \varepsilon\epsilon_n^2}$$
\end{proposition}

\noindent \textit{Proof:} Let $\mathcal{F}_{jn}=\{\theta_{jn}: |\theta_{jn}|\leq C_n\}$
	$$\mathcal{F}_n=\cap_{j=1}^{K(n)} \mathcal{F}_{jn}\implies \mathcal{F}_n^c= \cap_{j=1}^{K(n)}\mathcal{F}_{jn}^c$$
Note that
\begin{align*}
	\int_{\btheta_n \in \mathcal{F}_n^c}p(\btheta_n)d\btheta_n&\leq \sum_{j=1}^{K(n)}\int_{\mathcal{F}_{jn}^c}\frac{1}{\sqrt{2\pi \zeta_{jn}^2}}e^{-\frac{(\theta_{jn}-\mu_{jn})^2}{2\zeta_{jn}^2}}d\theta_{jn}\\ &=\sum_{j=1}^{K(n)}\int_{-\infty}^{-C_n}\frac{1}{\sqrt{2\pi \zeta_{jn}^2}}e^{-\frac{(\theta_{jn}-\mu_{jn})^2}{2\zeta_{jn}^2}}d\theta_{jn}+\sum_{j=1}^{K(n)}\int_{C_n}^{\infty}\frac{1}{\sqrt{2\pi \zeta_{jn}^2}}e^{-\frac{(\theta_{jn}-\mu_{jn})^2}{2\zeta_{jn}^2}}d\theta_{jn}\\
	&=\sum_{j=1}^{K(n)}\left(1-\Phi\left(\frac{C_n-\mu_{jn}}{\zeta_{jn}}\right)\right)+\sum_{j=1}^{K(n)}\left(1-\Phi\left(\frac{C_n+\mu_{jn}}{\zeta_{jn}}\right)\right)
\end{align*}

\noindent Since $||\bmu_n||_2^2=o(n\epsilon_n^2) \implies ||\bmu_n||_\infty=o(\sqrt{n}\epsilon_n)$. Also, $||\bzeta_n||_\infty=O(n)$, which implies for some $M>0$,
\begin{equation}
\label{e:c-lb}
    \min \left(\frac{|C_n-\mu_{jn}|}{\zeta_{jn}},\frac{|C_n+\mu_{jn}|}{\zeta_{jn}}\right)\geq \frac{(C_n-\sqrt{n})}{nM}\geq e^{\log C_n-2\log n}-\frac{1}{\sqrt{n}M} \sim e^{R_n\log n} \to \infty 
\end{equation}
where the last asymptotic relation holds because $1/\sqrt{n}\to 0$ and $R_n=(n^b \epsilon_n^2)/(K(n)\log n)-2 \to \infty$ since $K(n)\log n=o(n^b \epsilon_n^2)$. 

\noindent Thus, using Mill's ratio, we get:
\begin{align*}
	\int_{\btheta_n \in \mathcal{F}_n^c}p(\btheta_n)d\btheta_n&	\lesssim \sum_{j=1}^{K(n)}\frac{\zeta_{jn}}{C_n-\mu_{jn}}e^{-\frac{(C_n-\mu_{jn})^2}{2\zeta_{jn}^2}}+\sum_{j=1}^{K(n)}\frac{\zeta_{jn}}{C_n+\mu_{jn}}e^{-\frac{(C_n+\mu_{jn})^2}{2\zeta_{jn}^2}}\\
&\leq 2K(n)e^{-\frac{(C_n-\sqrt{n})^2}{2n^2M^2}}\lesssim e^{-\varepsilon n\epsilon_n^2}
\end{align*}
where the last asymptotic inequality holds because \begin{align*}
    \frac{(C_n-\sqrt{n})^2}{2n^2M^2}-\log 2K(n)\gtrsim \frac{1}{2}e^{2R_n\log n} -2\log n=n\left(\frac{e^{R_n}}{2}-\frac{2\log n}{n}\right) \geq \varepsilon n\epsilon_n^2
\end{align*}
In the above step, the first asymptotic inequality holds due to \eqref{e:c-lb} and $K(n)\leq n$. The last inequality holds since $R_n \to \infty$ and $\log /n \to 0$.
\begin{lemma}
\label{lem:v-bound}
Let $n\epsilon_n^2\to \infty$. Suppose $K(n)\log n =o(n^b\epsilon_n^2)$ for some $0<b<1$ and $p(\btheta_n)$ satisfies
 \eqref{e:prior} with $||\bmu_n||_2^2=o(n\epsilon_n^2)$. Then for every $\varepsilon>0$,
 $$\log \int_{\mathcal{U}_{\varepsilon \epsilon_n}^c} \frac{L(\btheta_n)}{L_0} p(\btheta_n)d\btheta_n\leq \log 2-\varepsilon^2 n\epsilon_n^2  + o_{P_0^n}(1)$$
 \end{lemma}

\noindent \textit{Proof.} In this direction, we first show
\begin{equation}
\label{e:hell-split}
    P_0^n\left( \int_{\mathcal{U}_{\varepsilon \epsilon_n}^c} \frac{L(\btheta_n)}{L_0} p(\btheta_n)d\btheta_n>  2e^{-\varepsilon n\epsilon_n^2}\right)\to 0,\:\: n \to \infty
\end{equation}
\begin{align*}
&	P_0^n\left(\int_{\mathcal{U}_{\varepsilon \epsilon_n}^c} \frac{L(\btheta_n)}{L_0} p(\btheta_n)d\btheta_n> 2e^{-\varepsilon^2 n\epsilon_n^2}\right)\\
	&\leq P_0^n\left(\int_{\mathcal{U}_{\varepsilon \epsilon_n}^c \cap \mathcal{F}_n} \frac{L(\btheta_n)}{L_0} p(\btheta_n)d\btheta_n> e^{-\varepsilon^2 n\epsilon_n^2}\right)+P_0^n\left(\int_{\mathcal{F}_n^c} \frac{L(\btheta_n)}{L_0} p(\btheta_n)d\btheta_n> e^{-\varepsilon^2 n\epsilon_n^2}\right)
\end{align*}
Using  lemma  \ref{lem:e-bound-0} with $\varepsilon=\varepsilon \epsilon_n$ and $C_n=e^{n^b\epsilon_n^2/K(n)}$, \begin{align*}\int_{\varepsilon^2\epsilon_n^2/8}^{{\sqrt{2}\varepsilon \epsilon_n}}{H_{[]}(u, \widetilde{\mathcal{F}}_n, ||.||_2)} du &\lesssim   \varepsilon \epsilon_n \sqrt{2K_n (\log K_n+2\log C_n-\log  \epsilon_n)}\\
&\leq \varepsilon \epsilon_n
O(\max(\sqrt{K(n)\log K(n)}, \sqrt{K(n)\log C_n}, \sqrt{-\log \epsilon_n}))\\
&\leq \varepsilon \epsilon_n \max(o(\sqrt{n}\epsilon_n), O(\sqrt{n^b} \epsilon_n),O(\sqrt{\log n}))\leq \varepsilon^2 \epsilon_n^2 \sqrt{n}
\end{align*}
where $H_{[]}(u,\widetilde{\mathcal{F}}_n,||.||_2)$ is as in definition \ref{def:hell-brack}. The first inequality in the third step follows because $K(n)\leq n$ and $K(n)\log n=o(n\epsilon_n^2)$,  $K(n)\log C_n =K(n)(n^b \epsilon_n^2 /K(n))$, $1/\epsilon_n^2=o(n) \implies -\log \epsilon_n^2\leq \log n$. The second inequality in the third step follows since $n^b/n=o(1)$ and  $\log n=o(n\epsilon_n^2)$.

\noindent By theorem 1 in \cite{WS1995}, for some constant $C>0$, we have
	\begin{align}
	    \label{e:p0-n-1}
 \nonumber P_0^n\left(\int_{\btheta_n\in \mathcal{U}_{\varepsilon\epsilon_n}^c \cap \mathcal{F}_n } \frac{L(\btheta_n)}{L_0 }p(\btheta_n)d\btheta_n> e^{-\varepsilon^2n\epsilon_n^2}\right)	&\leq P_0^n\left(\sup_{\btheta_n\in \mathcal{U}_{\varepsilon\epsilon_n}^c \cap \mathcal{F}_n } \frac{L(\btheta_n)}{L_0}> e^{-\varepsilon^2n\epsilon_n^2}\right)\\
 \leq 4\exp(-C\varepsilon^2 n\epsilon_n^2)\to 0\end{align}
\noindent Using proposition \ref{lem:p-bound-0} with $\varepsilon=2\varepsilon$, we have	$$\int_{\btheta_n \in \mathcal{F}_n^c} p(\btheta_n)d\btheta_n \leq e^{-2 n \varepsilon^2 {\epsilon_n}^2}$$
	Therefore, using lemma \ref{lem:pp-bound-0} with  $\varepsilon=2\varepsilon^2\epsilon_n^2$ and $\tilde{\varepsilon}={\varepsilon}^2 \epsilon_n^2$, we have
	
	\begin{equation} \label{e:p0-n-2}
	    P_0^n\left(\int_{\mathcal{F}_n^c} \frac{L(\btheta_n)}{L_0} p(\btheta_n)d\btheta_n> e^{-\varepsilon^2 n\epsilon_n^2}\right) \leq e^{-\varepsilon^2 n\epsilon_n^2} \to 0.
	    \end{equation}
Combining \eqref{e:p0-n-1} and \eqref{e:p0-n-2}, \eqref{e:hell-split} follows. 

\noindent Finally, to complete the proof, let  $\textcircled{1}=\log \int_{\mathcal{U}_{\varepsilon \epsilon_n}^c} (L(\btheta_n)/L_0) p(\btheta_n)d\btheta_n$.
\begin{align*}
    \textcircled{1}&=\textcircled{1} I_{\{\textcircled{1} \leq \log 2 -\varepsilon^2 \epsilon_n^2\}}+\textcircled{1} I_{\{\textcircled{1} > \log 2 -\varepsilon^2 \epsilon_n^2\} }\leq \log 2- \varepsilon^2 \epsilon_n^2+\underbrace{\textcircled{1} I_{\{\textcircled{1} > \log 2 -\varepsilon^2 \epsilon_n^2\} }}_{\textcircled{2}}\\
    &=\log 2- \varepsilon^2 \epsilon_n^2+o_{P_0^n}(1)
\end{align*}
where the last equality follows from \eqref{e:hell-split} as below
$$P_0^n(|\textcircled{2}|>\nu)\leq P_0^n(I_{\{\textcircled{1} > \log 2 -\varepsilon^2 \epsilon_n^2\}}=1) =P_0^n(\textcircled{1}> \log 2 -\varepsilon^2 \epsilon_n^2) \to 0.$$
\begin{proposition}
\label{lem:q-bound-v}
Let $p(\btheta_n)$ satisfy \eqref{e:prior} with $||\bzeta_n||=O(n)$ and $||\bzeta^*_n||_n=O(1)$, $\zeta^*_n=1/\bzeta_n$.
\begin{enumerate}
	\item If $K(n)\log n=o(n)$ and $||\bmu_n||_2^2=o(n)$, then
	\begin{equation}
	\label{e:q-bound-1}
	d_{\rm KL}(\pi^*,\pi(.|\boldy_n,\boldX_n))=o_{P_0^n}(n)
	\end{equation}
	\item If $K(n)\log n=o(n\epsilon_n^2)$ and $||\bmu_n||_2^2=o(n\epsilon_n^2)$ and there exists a neural network such $||\eta_0-\eta_{\boldt_n}||_\infty=o(n\epsilon_n^2)$ and $||\boldt_n||_2^2=o(n\epsilon_n^2)$, then
	\begin{equation}
	\label{e:q-bound-2}
	d_{\rm KL}(\pi^*,\pi(.|\boldy_n,\boldX_n))=o_{P_0^n}(n\epsilon_n^2)
	\end{equation}
\end{enumerate}
\end{proposition}

\noindent \textit{Proof.} For any $q \in \mathcal{Q}_n$. \begin{align}
\label{e:d-KL-break}
\nonumber	d_{\rm KL}(q,\pi(.|\boldy_n,\boldX_n))&=\int q(\btheta_n)\log q(\btheta_n)d\btheta_n-\int q(\btheta_n) \log \pi(\btheta_n|\boldy_n,\boldX_n)d\btheta_n\\
\nonumber	&=\int q(\btheta_n)\log q(\btheta_n)d\btheta_n- \int q(\btheta_n) \log \frac{L(\btheta_n)p(\btheta_n)}{\int L(\btheta_n)p(\btheta_n)d\btheta_n} d\btheta_n\\
\nonumber&=d_{\rm KL}(q,p)-\int \log  \frac{L(\btheta_n)}{L_0}q(\btheta_n) d\btheta_n+\log \int \frac{L(\btheta_n)}{L_0}  p(\btheta_n) d\btheta_n\\
&\leq d_{\rm KL}(q,p)+\left|\int \log  \frac{L(\btheta_n)}{L_0}q(\btheta_n) d\btheta_n\right|+\left|\log \int \frac{L(\btheta_n)}{L_0}  p(\btheta_n) d\btheta_n\right|
\end{align}
Since $\pi^*$ satisfies minimizes the KL-distance to $\pi(.|\boldy_n,\boldX_n)$ in the family $\mathcal{Q}_n$, therefore 
\begin{align}
\label{e:kl-2}
P_0^n\left(d_{\rm KL}(\pi^*,\pi(.|\boldy_n,\boldX_n))>\kappa\right) \leq P_0^n\left(d_{\rm KL}(q,\pi(.|\boldy_n,\boldX_n))>\kappa\right) 
\end{align}
for any $\kappa>0$.

\noindent \textit{Proof of part 1.}   Note, $K(n)\log n=o(n)$, $||\mu_n||_2^2 =o(n)$, $||\bzeta_n||_\infty=O(n)$ and $||\bzeta^*_n||_\infty=O(1)$. We take $q(\btheta_n)=MVN(\boldt_n, \boldI_{K(n)}/\sqrt{n})$ where $\boldt_n$ is defined next.

\noindent For $N\geq 1$, let $\eta_{\boldt_N}$ be a neural satisfying $||\eta_{\boldt_N}-\eta_0||_\infty \leq \varepsilon/4$.  The existence of such a neural network is always guaranteed by \cite{HOR}.  
Define $\boldt_n$ as $$\beta^{\boldt^n}_{j} =\begin{cases}
\beta^{\boldt^N}_{j},& j=1, \cdots, k_N\\
0,& j=k_N+1, \cdots, k_n
\end{cases} \hspace{5mm} \gamma^{\boldt^n}_{j} =\begin{cases}
\gamma^{\boldt^N}_{j},& j=1, \cdots, k_N\\
0,& j=k_N+1, \cdots, k_n
\end{cases} $$
The above choice guarantees $||\eta_{\boldt_n}-\eta_0||_\infty \leq \varepsilon/4$.

\noindent {\bf Step 1 (a):} Using proposition \ref{e:kl-q-p}, with $\epsilon_n=1$, we get for any $\nu>0$,
\begin{align*}
d_{\rm KL}(q,p)&\leq n\nu
\end{align*} 
where the above step follows $||\boldt_n||_2^2=||\boldt_N||_2^2$ is bounded which implies $||\boldt_n||_2^2=o(n)$. Therefore,
\begin{equation}
\label{e:kl-split-1}
P_0^n(d_{\rm KL}(q,p)>n\nu)=0
\end{equation}

\noindent {\bf Step 1 (b):} Next, note that
\begin{align}
	\label{e:dk-bound-1}
	\nonumber d_{\rm KL}(\ell_0,\ell_{\btheta_n})&= \int_{\boldx \in [0,1]^p} \left ( \sigma(\eta_0(\boldx))\log \frac{\sigma(\eta_0(\boldx))}{\sigma(\eta_{\btheta_n}(\boldx))}+(1-\sigma(\eta_0(\boldx)))\log \frac{1-\sigma(\eta_0(\boldx))}{1-\sigma(\eta_{\btheta_n}(\boldx))}\right)d\boldx\\
	&=\int_{\boldx \in [0,1]^p} \left(\sigma(\eta_0(\boldx))(\sigma(\eta_{\btheta_n}(\boldx))-\sigma(\eta_0(\boldx)))+\log \frac{1-\sigma(\eta_0(\boldx))}{1-\sigma(\eta_{\btheta_n}(\boldx))}\right)d\boldx
		\end{align}
Since $||\eta_0-\eta_{\boldt_n}||_\infty \leq \varepsilon/4$, using proposition \ref{lem:f-bound} with $\epsilon_n=1$ and $\varepsilon=\varepsilon$
\begin{align*}\int d_{\rm KL}(\ell_0,\ell_{\btheta_n})q(\btheta_n)d\btheta_n \leq \varepsilon
\end{align*}
where the above step follows since $||\boldt_n||_2^2=||\boldt_N||_2^2$ is bounded which implies $||\boldt_n||_2^2=o(n)$.

\noindent Therefore, by lemma \ref{lem:c-bound-0}, 
\begin{equation}
\label{e:kl-split-2}
P_0^n\left(\left|\int \log  \frac{L(\btheta_n)}{L_0}q(\btheta_n) d\btheta_n\right|> n\nu\right)\leq \frac{\varepsilon}{\nu}.
\end{equation}
{\bf Step 1 (c):} Since $||\eta_0-\eta_{\boldt_n}||_\infty\leq \varepsilon/4$, therefore using proposition \ref{lem:kl-bound-v} with $\epsilon_n=1$ and $\nu=\varepsilon$ we get
\begin{align*}
\int_{\btheta_n\in \mathcal{N}_\varepsilon} p(\btheta_n)d\btheta_n &
\geq \exp(-n\varepsilon)
\end{align*}
where the above step follows $||\boldt_n||_2^2=||\boldt_N||_2^2$ is bounded which implies $||\boldt_n||_2^2=o(n)$.

\noindent Therefore, using lemma \ref{lem:b-bound-0}, we get 
\begin{equation}
\label{e:kl-split-3}
P_0^n\left(\left|\log  \int \frac{L(\btheta_n)}{L_0}p(\btheta_n) d\btheta_n\right|> n\nu\right)\leq  \frac{2\varepsilon}{\nu}
\end{equation}
{\bf Step 1 (d):} From \eqref{e:kl-2} and \eqref{e:d-KL-break} we get
\begin{align}
\label{e:kl-split-4}
\nonumber P_0^n(d_{\rm KL}(\pi^*,\pi(.|\boldy_n,\boldX_n))>3n\nu)&\leq P_0^n \left(d_{\rm KL}(q,p)>n\nu\right)\\
&\hspace{-60mm}+P_0^n\left(\left|\int \log  \frac{L(\btheta_n)}{L_0}q(\btheta_n) d\btheta_n\right|> n\nu\right)+P_0^n\left(\left|\log \int   \frac{L(\btheta_n)}{L_0}p(\btheta_n) d\btheta_n\right|> n\nu\right)\leq \frac{3\varepsilon}{\nu}
\end{align}
where the last inequality is a consequence of \eqref{e:kl-split-1}, \eqref{e:kl-split-2} and \eqref{e:kl-split-3}. 

\noindent Since $\varepsilon$ is arbitrary, taking $\varepsilon \to 0$ completes the proof.

\noindent \textit{Proof of part 2.}   Note, $K(n)\log n=o(n\epsilon_n^2)$, $||\mu_n||_2^2 =o(n\epsilon_n^2)$, $||\bzeta_n||_\infty=O(n)$ and $||\bzeta^*_n||_\infty=O(1)$. We take $q(\btheta_n)=MVN(\boldt_n, \boldI_{K(n)}/\sqrt{n})$ where $\boldt_n$ is defined next.

\noindent Let $\eta_{\boldt_n}$ be the neural satisfying 
\begin{align*}
||\eta_{\boldt_n}-\eta_0||_\infty\leq \varepsilon\epsilon_n^2/4 \hspace{5mm} ||\boldt_n||_2^2=o(n\epsilon_n^2)
\end{align*}
The existence of such a neural network is guaranteed since $||\eta_{\boldt_n}-\eta_0||_\infty=o(\epsilon_n^2)$.

\noindent {\bf Step 2 (a):} Since $||\boldt_n||_2^2=o(n\epsilon_n^2)$, by proposition \ref{e:kl-q-p}, 
\begin{align*}
d_{\rm KL}(q,p)&\leq 
n \epsilon_n^2\nu
\end{align*} 
which implies
\begin{equation}
\label{e:kl-split-5}
P_0^n(d_{\rm KL}(q,p)>\nu n\epsilon_n^2)=0
\end{equation}

\noindent {\bf Step 2 (b):}  Since $||\eta_{\boldt_n}-\eta_0||_\infty\leq \varepsilon\epsilon_n^2/4$ and $||\boldt_n||_2^2=o(n\epsilon_n^2)$, by proposition \ref{lem:f-bound}, \begin{align*}\int d_{\rm KL}(\ell_0,\ell_{\btheta_n})q(\btheta_n)d\btheta_n
\leq \varepsilon \epsilon_n^2
\end{align*}
\noindent Therefore, by lemma \ref{lem:c-bound-0}, 

\begin{equation}
\label{e:kl-split-6}
P_0^n\left(\left|\int \log  \frac{L(\btheta_n)}{L_0}q(\btheta_n) d\btheta_n\right|> \nu n \epsilon_n^2\right)\leq \frac{\varepsilon }{\nu}.
\end{equation}

\noindent {\bf Step 2 (c):}  Since $||\eta_0-\eta_{\boldt_n}||_\infty\leq \varepsilon\epsilon_n^2/4$ and $||\boldt_n||_2^2=o(n\epsilon_n^2)$, by proposition \ref{lem:kl-bound-v}, 
\begin{align*}
\int_{\btheta_n\in \mathcal{N}{\varepsilon \epsilon_n^2}} p(\btheta_n)d\btheta_n 
&\geq \exp(-\varepsilon n\epsilon_n^2)
\end{align*}
\noindent Therefore, using lemma \ref{lem:b-bound-0}, we get 
\begin{equation}
\label{e:kl-split-7}
P_0^n\left(\left|\log  \int \frac{L(\btheta_n)}{L_0}q(\btheta_n) d\btheta_n\right|> \nu n \epsilon_n^2\right)\leq  \frac{2\varepsilon}{\nu}
\end{equation}

\noindent {\bf Step 2 (d):} From \eqref{e:kl-2} and \eqref{e:d-KL-break} we get
\begin{align}
\label{e:kl-split-4}
\nonumber P_0^n(d_{\rm KL}(\pi^*,\pi(.|\boldy_n,\boldX_n))>3\nu n\epsilon_n^2)&\leq P_0^n \left(d_{\rm KL}(q,p)>\nu n\epsilon_n^2\right)\\
&\hspace{-60mm}+P_0^n\left(\left|\int \log  \frac{L(\btheta_n)}{L_0}q(\btheta_n) d\btheta_n\right|> \nu n\epsilon_n^2\right)+P_0^n\left(\left|\log \int   \frac{L(\btheta_n)}{L_0}p(\btheta_n) d\btheta_n\right|> \nu n\epsilon_n^2\right)\leq \frac{3\varepsilon}{\nu}
\end{align}
where the last inequality is a consequence of \eqref{e:kl-split-5}, \eqref{e:kl-split-6} and \eqref{e:kl-split-7}. 

\noindent Since $\varepsilon$ is arbitrary, taking $\varepsilon \to 0$ completes the proof.

\subsection{Theorems and Corollaries}

\noindent Let $\mathcal{U}_{\varepsilon}$ be as in \eqref{e:hell-def}. Let  $\pi(.|\boldy_n,\boldX_n)$ and $\pi^*$ be the posterior and the variational posterior as in \eqref{e:posterior} and \eqref{e:var-posterior} respectively.

\noindent {\bf Proof of Theorem \ref{thm:post-cons}:}

\noindent We assume Relation \eqref{e:kl-ub-lb} holds with $A_n$ and $B_n$ are same as in \eqref{e:A-B-def}.

\noindent By assumptions (A1) and (A2), the prior parameters satisfy $$||\bmu_n||_2^2=o(n), \:\:||\bzeta_n||_\infty=O(n),\:\:||\bzeta_n^*||_\infty=O(1), \:\: \bzeta^*_n=1/\bzeta_n.$$

\noindent Note  $K(n) \sim k_n\sim n^a$, $0<a<1$ which implies $K(n)\log n=o(n)$.

\noindent By proposition \ref{lem:q-bound-v} part 1., \begin{equation}
\label{e:kl-val-1}
    d_{\rm KL} (\pi^*, \pi(.|\boldy_n,\boldX_n))= o_{P_0^n}(n).
\end{equation}
\noindent By step 1 (c) in the proof of proposition \ref{lem:q-bound-v}
\begin{equation}
\label{e:kl-val-2}
B_n= o_{P_0^n}(n)
\end{equation}
\noindent Since, $K(n) \sim n^a$, $K(n)\log n=o(n^b)$, $a<b<1$. Using proposition \ref{lem:v-bound} with $\epsilon_n=1$, 
\begin{equation}
\label{e:kl-val-3}
-\pi^*(\mathcal{U}_\varepsilon^c)A_n \geq n\varepsilon^2\pi^*(\mathcal{U}_\varepsilon^c)-\log 2+o_{P_0^n}(1)=n\varepsilon^2\pi^*(\mathcal{U}_\varepsilon^c)+O_{P_0^n}(1)
\end{equation}

\noindent Thus, using \eqref{e:kl-val-1}, \eqref{e:kl-val-2} and \eqref{e:kl-val-3} in \eqref{e:kl-ub-lb}, we get
$$n\varepsilon^2\pi^*(\mathcal{U}_\varepsilon^c)+O_{P_0^n}(1)\leq o_{P_0^n}(n)+o_{P_0^n}(n) \implies \pi^*(\mathcal{U}_\varepsilon^c)=o_{P_0^n}(1)$$

\noindent {\bf Proof of Theorem \ref{thm:post-cons-del}:}

\noindent We assume Relation \eqref{e:kl-ub-lb} holds with $A_n$ and $B_n$ are same as in \eqref{e:A-B-def}.

\noindent Let  $k_n\sim n^a$ and $\epsilon_n^2\sim n^{-\delta}$, $0<\delta<1-a$. This implies $K(n)\log n=o(n\epsilon_n^2)$. 

\noindent By assumptions (A1) and (A4), the prior parameters satisfy $$||\bmu_n||_2^2=o(n \epsilon_n^2), \:\:||\bzeta_n||_\infty=O(n),\:\:||\bzeta_n^*||_\infty=O(1),  \:\: \bzeta^*_n=1/\bzeta_n.$$
Also by assumption (A3),
$$||\eta_0-\eta_{\boldt_n}||_\infty=o(\epsilon_n^2), \:\: ||\boldt_n||_2^2=o(n\epsilon_n^2)$$
\noindent By proposition \ref{lem:q-bound-v} part 2., \begin{equation}
\label{e:kl-val-4}
    d_{\rm KL} (\pi^*, \pi(.|\boldy_n,\boldX_n))= o_{P_0^n}(n\epsilon_n^2).
\end{equation}
\noindent By step 2 (c) in the proof of proposition \ref{lem:q-bound-v}
\begin{equation}
\label{e:kl-val-5}
    B_n= o_{P_0^n}(n \epsilon_n^2 )
\end{equation}
\noindent Since $K(n) \sim n^a$, $K(n)\log n=o(n^b \epsilon_n^2)$, $a+\delta<b<1$. Using proposition \ref{lem:v-bound}, it follows that 
\begin{equation}
\label{e:kl-val-6}
-\pi^*(\mathcal{U}_{\varepsilon\epsilon_n}^c )A_n \geq \varepsilon^2 n \epsilon_n^2 \pi^*(\mathcal{U}_{\varepsilon\epsilon_n}^c)-\log 2+o_{P_0^n}(1)=\varepsilon^2 n \epsilon_n^2 \pi^*(\mathcal{U}_{\varepsilon\epsilon_n}^c)+O_{P_0^n}(1)
\end{equation}

\noindent Thus, using \eqref{e:kl-val-4}, \eqref{e:kl-val-5} and \eqref{e:kl-val-6} in \eqref{e:kl-ub-lb}, we get
$$n\varepsilon^2 \epsilon_n^2\pi^*(\mathcal{U}_{\varepsilon\epsilon_n}^c)+O_{P_0^n}(1)\leq o_{P_0^n}(n\epsilon_n^2)+o_{P_0^n}(n\epsilon_n^2) \implies \pi^*(\mathcal{U}_{\varepsilon\epsilon_n}^c)=o_{P_0^n}(1)$$

\noindent {\bf Proof of Corollary 1.}
	
\noindent Let $\hat{\ell}_n(y,\boldx)=\int \ell_{\btheta_n}(y,\boldx) \pi^*(\btheta_n)d\btheta_n$.
	\begin{align*}
	d_{\rm H}(\hat{\ell}_n,\ell_0)&=d_{\rm H}\left( \int \ell_{\btheta_n} \pi^*(\btheta_n)d\btheta_n,\ell_0\right)\\
	&\leq \int d_{\text{H}}(\ell_{\btheta_n},\ell_0) \pi^*(\btheta_n)d\btheta_n \hspace{5mm} \text{Jensen's inequality}\\
	&=\int_{\mathcal{U}_\varepsilon} d_{\text{H}}(\ell_{\btheta_n},\ell_0) \pi^*(\btheta_n)d\btheta_n+\int_{\mathcal{U}_\varepsilon^c} d_{\text{H}}(\ell_{\btheta_n},\ell_0) \pi^*(\btheta_n)d\btheta_n\\
	&\leq \varepsilon+o_{P_0^n}(1)
	\end{align*}
	Taking $\varepsilon \to 0$, we get $d_{\text{H}}(\hat{\ell}_n,\ell_0)=o_{P_0^n}(1)$.
	
\noindent By \eqref{e:eta-hat-def}, note that $\hat{\eta}(y,\boldx)=\sigma^{-1} \hat{\ell}_n(y,\boldx)=\log \frac{ \hat{\ell}_n(1,\boldx)}{\hat{\ell}_n(0,\boldx)}$, then
\begin{align}
	\label{e:hell-lik}
\nonumber	2d^2_{\rm H}(\hat{\ell}_n,\ell_0)&=\int_{\boldx \in [0,1]^p} \sum_{y \in \{0,1\}} \left(\sqrt{\hat{\ell}_n(y,\boldx)}-\sqrt{\ell_0(y,\boldx)}\right)^2 d\boldx\\
\nonumber	&=2-2 \int_{\boldx \in [0,1]^p} \sum_{y \in \{0,1\}}  \sqrt{\hat{\ell}_n(y,\boldx)\ell_0(y,\boldx)} d\boldx\\
\nonumber	&=2-2\int_{\boldx \in [0,1]^p} \sum_{y \in \{0,1\}} e^{\left\{\frac{1}{2}\left(y\hat{\eta}(\boldx)-\log(1+e^{\hat{\eta}(\boldx)})+y\eta_0(\boldx)-\log(1+e^{\eta_0(\boldx)}\right)\right\}} d\boldx\\
\nonumber	&=2-2\int_{\boldx \in [0,1]^p} \left(\sqrt{\sigma(\eta_0(\boldx))\sigma(\hat{\eta}(\boldx))}+\sqrt{(1-\sigma(\eta_0(\boldx)))(1-\sigma(\hat{\eta}(\boldx)))}\right) d\boldx\\
\nonumber	&\geq 2-2 \int_{\boldx \in [0,1]^p} \sqrt{1- (\sqrt{\sigma(\eta_0(\boldx))}-\sqrt{\sigma(\hat{\eta}(\boldx))})^2}d\boldx\\
\nonumber	&\geq \int_{\boldx \in [0,1]^p} (\sqrt{\sigma(\eta_0(\boldx))}-\sqrt{\sigma(\hat{\eta}(\boldx))})^2d\boldx \\
& \geq \frac{1}{4}\int_{\boldx \in [0,1]^p} (\sigma(\eta_0(\boldx))-\sigma(\hat{\eta}(\boldx)))^2d\boldx 
	\end{align}
In the above equation,	the sixth and the seventh step hold because $\sqrt{1-x}\leq 1-x/2$ and $|p_1-p_2|\leq |\sqrt{p_1}+\sqrt{p_2}||\sqrt{p_1}-\sqrt{p_2}|\leq 2|\sqrt{p_1}-\sqrt{p_2}|$ repsectively. The fifth step holds because
	\begin{align*}
	    \left(\sqrt{p_1p_2}+\sqrt{(1-p_1)(1-p_2)}\right)^2 &=p_1p_2+1-p_1-p_2+\sqrt{p_1p_2(1-p_1)(1-p_2)}\\
	    &\leq \sqrt{p_1p_2}+1-p_1-p_2+\sqrt{p_1p_2}=1-(\sqrt{p1}-\sqrt{p_2})^2
	\end{align*}
By \eqref{e:hell-lik} and Cauchy Schwartz inequality,  \vspace{-4mm}
\begin{align}
    \label{e:hell-upp}
\nonumber    \int_{\boldx \in [0,1]^p} |\sigma(\eta_0(\boldx))-\sigma(\hat{\eta}(\boldx))| d\boldx &\leq \left(\int_{\boldx \in [0,1]^p}  (\sigma(\eta_0(\boldx))-\sigma(\hat{\eta}(\boldx)))^2d\boldx\right)^{1/2}\\
    &\leq 2\sqrt{2} d_{\text{H}}(\hat{\ell}_n, \ell_0)=o_{P_0^n}(1)
\end{align}
The proof follows in lieu of \eqref{e:r-c-eta}.

\noindent {\bf Proof of Corollary 2.}

\noindent We assume Relation \eqref{e:kl-ub-lb} holds with $A_n$ and $B_n$ are same as in \eqref{e:A-B-def}.

\noindent Let  $k_n\sim n^a$ and $\epsilon_n^2\sim n^{-\delta}$, $0<\delta<1-a$. This implies $K(n)\log n=o(n\epsilon_n^2)$. 

\noindent Also $K(n)\log n=o(n^b \epsilon_n^2)$, $a+\delta<b<1$. This implies $K(n)\log n =o(n^b (\epsilon_n^2)^\kappa)$, $0\leq \kappa \leq 1$.
Thus, using proposition \ref{lem:v-bound} with $\epsilon_n=\epsilon_n^{k}$, we get
\begin{equation}
-\pi^*(\mathcal{U}_{\varepsilon\epsilon_n^{\kappa}}^c )A_n \geq \varepsilon^2 n \epsilon_n^{2\kappa} \pi^*(\mathcal{U}_{\varepsilon\epsilon_n^{\kappa}}^c)-\log 2+o_{P_0^n}(1)=\varepsilon^2 n \epsilon_n^{2\kappa} \pi^*(\mathcal{U}_{\varepsilon\epsilon_n^k}^c)+O_{P_0^n}(1)
\end{equation}
This together with \eqref{e:kl-val-4}, \eqref{e:kl-val-5} and \eqref{e:kl-ub-lb} implies 
$$\pi^*(\mathcal{U}_{\varepsilon\epsilon_n^\kappa}^c)=o_{P_0^n}(\epsilon_n^{2-2\kappa})$$

\noindent Let $\hat{\ell}_n(y,\boldx)=\int \ell_{\btheta_n}(y,\boldx) \pi^*(\btheta_n)d\btheta_n$.
	\begin{align*}
	d_{\rm H}(\hat{\ell}_n,\ell_0)&\leq \int_{\mathcal{U}_{\varepsilon \epsilon_n^\kappa}} d_{\text{H}}(\ell_{\btheta_n},\ell_0) \pi^*(\btheta_n)d\btheta_n+\int_{\mathcal{U}_{\varepsilon \epsilon_n^\kappa}^c} d_{\text{H}}(\ell_{\btheta_n},\ell_0) \pi^*(\btheta_n)d\btheta_n\\
	&\leq \varepsilon \epsilon_n^\kappa+o_{P_0^n}(\epsilon_n^{2-2\kappa})\\
&\hspace{-40mm}\text{Dividing  by $\epsilon_n^\kappa$ on both sides we get}\\
	\frac{1}{\epsilon_n^\kappa}d_{\rm H}(\hat{\ell}_n,\ell_0)&=o_{P_0^n}(\epsilon_n^{2-3\kappa})+o_{P_0^n}(1)=o_{P_0^n}(1), \hspace{5mm}0\leq \kappa\leq 2/3.
	\end{align*}
By \eqref{e:hell-upp}, for every $0\leq \kappa \leq 2/3$, $$\frac{1}{\epsilon_n^\kappa}\int_{\boldx \in U[0,1]^p} |\sigma(\eta_0(\boldx))-\sigma(\hat{\eta}(\boldx))| d\boldx\leq \frac{1}{\epsilon_n^\kappa}2\sqrt{2}d_{\text{H}}(\hat{\ell}_n,\ell_0)=o_{P_0^n}(1).$$
The proof follows in lieu of \eqref{e:r-c-eta}.

\section*{(C) Consistency of the true posterior.}

\begin{theorem}
\label{thm:true-post-cons}
Suppose  $k_n \sim n^a$, $0<a<1$. Additionally, the prior parameters in \eqref{e:prior} satisfies assumption (A1) and (A2). Then, 
\begin{enumerate}
    \item $$P_0^n\left(\pi(\mathcal{U}_{\varepsilon}^c|\boldy_n,\boldX_n)\leq 2e^{-n\varepsilon^2/2}\right) \to 1, n \to \infty$$
\item $$P_0^n(|R(\hat{C})-R(C^{\rm Bayes})|\leq 4\sqrt{2}\varepsilon)\to 1,n \to \infty $$
\end{enumerate}
where $R$ is the risk defined in \eqref{e:riskintegral}.
\end{theorem}

\noindent \textit{Proof.} From \eqref{e:posterior}, note that
\begin{align}
\label{e:post-break}
 \nonumber   \pi(\mathcal{U}_{\varepsilon}^c|\boldy_n,\boldX_n)&=\frac{\int_{\mathcal{U}_\varepsilon^c}L(\btheta_n)p(\btheta_n)d\btheta_n}{\int L(\btheta_n)p(\btheta_n)d\btheta_n}\\
    &=\frac{\int_{\mathcal{U}_\varepsilon^c}(L(\btheta_n)/L_0)p(\btheta_n)d\btheta_n}{\int (L(\btheta_n)/L_0)p(\btheta_n)d\btheta_n}
\end{align}
\noindent By assumptions (A1) and (A2), the prior parameters satisfy $$||\bmu_n||_2^2=o(n), \:\:||\bzeta_n||_\infty=O(n),\:\:||\bzeta_n^*||_\infty=O(1), \:\: \bzeta^*_n=1/\bzeta_n.$$

\noindent Note  $K(n) \sim k_n\sim n^a$, $0<a<1$ which implies $K(n)\log n=o(n)$. Thus, the conditions of proposition \ref{lem:kl-bound-v} hold with $\epsilon_n=1$. 
\begin{align}
\label{e:post-cons-1}
P_0^n\left(  \int \frac{L(\btheta_n)}{L_0}p(\btheta_n) d\btheta_n \leq e^{-n\nu}\right)&\leq P_0^n\left(\left|\log  \int \frac{L(\btheta_n)}{L_0}p(\btheta_n) d\btheta_n\right|> n\nu\right) \to 0, n \to \infty
\end{align}
where the above convergence follows from \eqref{e:kl-split-3} in step 1 (c) in the proof of proposition \ref{lem:q-bound-v}.

\noindent Also, since $K(n)\log n=o(n^b)$, $a<b<1$. Thus, conditions of proposition \ref{lem:v-bound} hold with $\epsilon_n=1$. 
\begin{align}
\label{e:post-cons-2}
P_0^n\left(  \int_{\mathcal{U}_\varepsilon^c} \frac{L(\btheta_n)}{L_0}p(\btheta_n) d\btheta_n \geq 2 e^{-n\varepsilon^2 }\right)\to 0, n \to \infty
\end{align}
where the last equality follows from \eqref{e:hell-split} with $\epsilon_n=1$ in the proof of Propostion \ref{lem:v-bound}.

Using \eqref{e:post-cons-1} and \eqref{e:post-cons-2} with \eqref{e:post-break}, we get
$$P_0^n\left(\pi(\mathcal{U}_{\varepsilon}^c|\boldy_n,\boldX_n)\geq 2e^{-n(\varepsilon^2-\nu)}\right) \to 0, n \to \infty$$
Take $\nu=\varepsilon^2/2$ to complete the proof.

Let $\hat{\ell}_n(y,\boldx)=\int \ell_{\btheta_n}(y,\boldx) \pi(\btheta_n|\boldy_n,\boldX_n)d\btheta_n$.
Mimicking the steps in the proof of corollary \ref{cor:fun-cons}, we get
	\begin{align*}
	d_{\rm H}(\hat{\ell}_n,\ell_0)&=d_{\rm H}\left( \int \ell_{\btheta_n} \pi(\btheta_n|\boldy_n,\boldX_n)d\btheta_n,\ell_0\right)\\
	&\leq \int d_{\text{H}}(\ell_{\btheta_n},\ell_0) \pi(\btheta_n|\boldy_n,\boldX_n)d\btheta_n \hspace{5mm} \text{Jensen's inequality}\\
	&=\int_{\mathcal{U}_\varepsilon} d_{\text{H}}(\ell_{\btheta_n},\ell_0) \pi(\btheta_n|\boldy_n,\boldX_n)d\btheta_n+\int_{\mathcal{U}_\varepsilon^c} d_{\text{H}}(\ell_{\btheta_n},\ell_0) \pi(\btheta_n|\boldy_n,\boldX_n)d\btheta_n\\
	&\leq \varepsilon+2e^{-n\varepsilon^2/2}\leq 2\varepsilon,\:\: \text{with probability tending to 1 as $n\to \infty$}\end{align*}
where the second last inequality is a consequence of part 1. in theorem \ref{thm:true-post-cons}. 
The remaining part of the proof follows by \eqref{e:hell-upp} and \eqref{e:r-c-eta}.

\begin{theorem}
\label{thm:true-post-cons-1}
Suppose  $k_n \sim n^a$, $0<a<1$, $\epsilon_n^2\sim n^{-\delta}$, $0<\delta<1-a$. Additionally, the prior parameters in \eqref{e:prior} satisfies assumption (A1) and (A4). Then, 
\begin{enumerate}
    \item 
$$P_0^n\left(\pi(\mathcal{U}_{\varepsilon \epsilon_n}^c|\boldy_n,\boldX_n)\leq 2e^{-n\epsilon_n^2\varepsilon^2/2}\right) \to 1, n \to \infty$$
\item $$P_0^n(|R(\hat{C})-R(C^{\rm Bayes})|\leq 4\sqrt{2}\varepsilon \epsilon_n)\to 1,n \to \infty $$
\end{enumerate}

\end{theorem}

\noindent \textit{Proof}. By assumptions (A1) and (A4), the prior parameters satisfy $$||\bmu_n||_2^2=o(n \epsilon_n^2), \:\:||\bzeta_n||_\infty=O(n),\:\:||\bzeta_n^*||_\infty=O(1),  \:\: \bzeta^*_n=1/\bzeta_n.$$
Also by assumption (A3),
$$||\eta_0-\eta_{\boldt_n}||_\infty=o(\epsilon_n^2), \:\: ||\boldt_n||_2^2=o(n\epsilon_n^2)$$
\noindent Note  $K(n) \sim k_n\sim n^a$, $0<a<1$ and $\epsilon_n\sim n^{-\delta}$, $0<\delta<1-a$, thus $K(n)\log n=o(n\epsilon_n^2)$. Thus, the conditions of proposition \ref{lem:kl-bound-v} hold.
\begin{align}
\label{e:post-cons-3}
P_0^n\left(  \int \frac{L(\btheta_n)}{L_0}p(\btheta_n) d\btheta_n \leq e^{-n\epsilon_n^2 \nu}\right)&\leq P_0^n\left(\left|\log  \int \frac{L(\btheta_n)}{L_0}p(\btheta_n) d\btheta_n\right|> n\epsilon_n^2\nu\right) \to 0, n \to \infty
\end{align}
where the above convergence follows from \eqref{e:kl-split-7} in step 2 (c) in the proof of proposition \ref{lem:q-bound-v}.

\noindent Also, since $K(n)\log n=o(n^b \epsilon_n^2)$, $a+\delta<b<1$. Thus conditions of proposition \ref{lem:v-bound} hold. 
\begin{align}
\label{e:post-cons-4}
P_0^n\left(  \int_{\mathcal{U}_{\varepsilon\epsilon_n}^c} \frac{L(\btheta_n)}{L_0}p(\btheta_n) d\btheta_n \geq 2 e^{-n \epsilon_n^2\varepsilon^2 }\right)\to 0, n \to \infty
\end{align}
where the last equality follows from \eqref{e:hell-split} in the proof of proposition \ref{lem:v-bound}.

Using \eqref{e:post-cons-3} and \eqref{e:post-cons-4} with \eqref{e:post-break}, we get
$$P_0^n\left(\pi(\mathcal{U}_{\varepsilon \epsilon_n}^c|\boldy_n,\boldX_n)\geq 2e^{-n\epsilon_n^2(\varepsilon^2-\nu)}\right) \to 0, n \to \infty$$
Take $\nu=\varepsilon^2/2$ to complete the proof.

Let $\hat{\ell}_n(y,\boldx)=\int \ell_{\btheta_n}(y,\boldx) \pi(\btheta_n|\boldy_n,\boldX_n)d\btheta_n$.
Mimicking the steps in the proof of corollary \ref{cor:fun-cons}, we get
	\begin{align*}
	d_{\rm H}(\hat{\ell}_n,\ell_0)&=d_{\rm H}\left( \int \ell_{\btheta_n} \pi(\btheta_n|\boldy_n,\boldX_n)d\btheta_n,\ell_0\right)\\
	&\leq \int d_{\text{H}}(\ell_{\btheta_n},\ell_0) \pi(\btheta_n|\boldy_n,\boldX_n)d\btheta_n \hspace{5mm} \text{Jensen's inequality}\\
	&=\int_{\mathcal{U}_{\varepsilon \epsilon_n}} d_{\text{H}}(\ell_{\btheta_n},\ell_0) \pi(\btheta_n|\boldy_n,\boldX_n)d\btheta_n+\int_{\mathcal{U}_{\varepsilon\epsilon_n}^c} d_{\text{H}}(\ell_{\btheta_n},\ell_0) \pi(\btheta_n|\boldy_n,\boldX_n)d\btheta_n\\
	&\leq \varepsilon\epsilon_n+2e^{-2n\epsilon_n^2\varepsilon^2}\leq 2\varepsilon\epsilon_n, \:\:\text{with probability tending to 1 as $n\to \infty$}	\end{align*}
where the second last inequality is a consequence of part 1. in theorem \ref{thm:true-post-cons-1} and the last inequality last equality follows since $\epsilon_n \sim n^{-\delta}$. Dividing by $\epsilon_n$ on both sides we get $$\epsilon_n^{-1}d_{\rm H}(\hat{\ell}_n,\ell_0) \leq 2\varepsilon,\:\:\text{with probability tending to 1 as $n\to \infty$}$$
The remaining part of the proof follows by \eqref{e:hell-upp} and \eqref{e:r-c-eta}.

\section*{(D) Tables}
\begin{table}[H]
\centering
\scalebox{1}{
\begin{tabular}{| l  c  c  c  c |}
\hline
Characteristics& MCI-S & MCI-C & Test statistic & P-value \\ \hline
Age(years) & $74.34\pm 7.78 $ & $74.84\pm 6.83 $  & -0.528 & $>0.5^a$\\
\hline 
Education(years) & $15.57\pm2.94  $ & $15.73 \pm 2.91 $  & -0.527 &$>0.5^b$ \\\hline 
APOE4 $\%$ & $34.65\%$ & $62.19\%$ & 17.900 &$<0.001^a$\\\hline

CDRSB  & $1.23 \pm 0.61 $ & $1.72 \pm 0.92$ & -5.237 &$<0.001^a$\\\hline

MMSE & $27.61 \pm1.74  $& $26.82\pm 1.71  $ &  3.645 &$<0.001^a$\\\hline

ADAS11 & $8.89 \pm3.79  $& $12.29 \pm 4.16 $ &  -6.823 &$<0.001^a$\\\hline

ADAS13 & $14.48 \pm 5.50 $& $20.01 \pm 5.79  $ &  -7.795 &$<0.001^a$\\\hline

ASASQ4 & $4.76 \pm 2.19  $& $6.77 \pm 2.21 $ &  -7.339 &$<0.001^a$\\\hline

RAVLT1 & $36.21\pm 10.10 $& $29.10 \pm 7.98 $ &  6.021 &$<0.001^a$\\\hline

RAVLT2 & $4.19\pm2.47  $& $2.91 \pm 2.26  $ &  4.231 &$<0.001^a$\\\hline

RAVLT3  & $4.31\pm2.59  $& $4.47 \pm 2.15 $ &  -1.501 &$0.135^a$\\\hline

RAVLT4  & $51.55 \pm31.04  $& $72.85 \pm 30.45 $ &   -5.464 &$<0.001^a$ \\\hline

LEDLTOTAL  & $4.96 \pm2.36 $& $3.41 \pm 2.66 $ &  4.931 &$<0.001^a$\\\hline

DIGTSCOR & $40.75 \pm 11.09 $& $36.72 \pm 10.96 $ &  2.883 &$<0.005^a$\\\hline

TRABSCOR & $109.43\pm62.94 $& $132.09\pm 71.36 $ &  -2.704 &$0.007^a$\\\hline

FAQ & $1.50 \pm2.99 $& $4.96 \pm 4.79$ & -7.243 &$<0.001^a$ \\\hline

mPACCdigit & $-5.376\pm2.96  $& $-8.06\pm 2.96 $ &  7.174 &$<0.001^a$\\\hline

mPACCtrailsB & $-5.47 \pm3.06$& $-8.22 \pm2.98  $ & 7.174 &$<0.001^a$\\\hline
\end{tabular}
}
\caption{{\bf Clinical Features and Cognitive Assessment Score.} Values are shown as mean $\pm$ standard deviation or percentage. Test statistics and P-values for differences between MCI-S and MCI-C are based on (a) t-test or (b) chi- square test. MCI-S = non-progressive MCI; MCI-P = progressive MCI; APOE = apolipoprotein E; MMSE = Mini-Mental State Examination. RAVLT = The Rey Auditory Verbal Learning Test (immediate: sum of 5 trails; learning: trial 5-trial 1; Forgetting: trial 5-delayed; perc.forgetting: Precent forgetting) ; DIGT = The Digit- Symbol Coding test; TRAB = Trail Making tests; CDRSB = Clinical Dementia Rating Scaled Response; FAQ = Activities of Daily living Score; ADAS = Alzheimer's Disease Assessment Scale–Cognitive sub- scale; mPACCdigit = the Digit Symbol Substitution Test from the Preclinical Alzheimer Cognitive Composite;}
\label{table:clinical_predictor}
\end{table}

\begin{table}[!h]
\centering
\scalebox{1}{
\renewcommand{\arraystretch}{1.1}%
\begin{tabular}{|ccccc|}
\hline
Characteristics& MCI-S & MCI-C &Test statistic &P-value\\ \hline 
HippoR & $3684 \pm 438 $ & $ 3366 \pm 437$ &  5.735&  $<0.001$\\
\hline
HippoL & $3414 \pm 418 $ & $3105 \pm 388$  &  5.994&  $<0.001$\\
\hline
flWMR & $96720 \pm 6218$ & $96976 \pm 5585$   &  -0.338  & 0.73\\
\hline
flWML & $93671 \pm 5836$ & $94238 \pm 5160 $&  -0.802  & 0.42\\
\hline
plWML & $50149 \pm 3714 $& $50038 \pm 3467$ &  0.242& 0.81\\
\hline
tlWMR  & $56076 \pm 3252 $& $55934 \pm 2931 $ & 0.359& 0.72\\
\hline

ACgCR  & $3167 \pm 756 $& $3128 \pm 641 $ & 0.438& 0.66\\
\hline
ACgCL  & $4104 \pm 787 $& $4075 \pm 689 $ & 0.312& 0.76 \\
\hline
EntR  & $2189 \pm 365 $& $1983 \pm 373 $ &  4.412& $<0.001$\\
\hline
EntL  & $2050 \pm 399 $& $1844 \pm 356 $ &  4.240& $<0.001$\\
\hline
MCgCR  & $4176 \pm 547 $& $4200 \pm 541 $ & -0.341& 0.73\\
\hline
MCgCL  & $3988 \pm 493 $& $4002 \pm 559 $ & -0.213& 0.83\\
\hline
MFCR  & $1581 \pm 342 $& $1505 \pm 524 $ & 1.805& 0.07\\
\hline
MFCL  & $1566 \pm 285 $& $1548 \pm 291 $ & 0.487& 0.62\\
\hline
OpIFGR  & $2575 \pm 608 $& $2425\pm 546 $ & 2.021& 0.04\\
\hline
OpIFGL  & $2465 \pm 550 $& $2361 \pm 579 $ &  1.466&0.14\\
\hline
OrIFGR  & $1252 \pm 315 $& $1196 \pm 362 $ &  1.322& 0.18\\
\hline
OrIFGL  & $1514 \pm 335 $& $1398 \pm 356 $ &  2.658& $<0.001$\\
\hline
PCgCR  & $3679 \pm 466 $& $3528 \pm 415 $ &  2.657&$<0.001$\\
\hline
PCgCL & $3991 \pm 442 $& $3789 \pm424 $ &  3.676& $<0.001$\\
\hline
PCuR  & $10129 \pm 1193 $& $9862 \pm 1313 $ & 1.701& 0.09\\
\hline
PCuL  & $10005 \pm1263 $& $9759 \pm 1299 $ & 1.522& 0.13\\
\hline
SPLR  & $8867 \pm1140 $& $8693 \pm 1219 $ & 1.180& 0.02\\
\hline
SPLL  & $8880 \pm1192 $& $8662 \pm 1313 $ & 1.390& 0.17\\

\hline
\end{tabular}
}
\caption{{\bf Significant MRI Features.} Values are shown as mean ± standard deviation or percentage. Test statistics and P-values for differences between MCI-C and MCI-S are based on  t-test. MCI-S = non-progressive MCI; MCI-C = progressive MCI. HippoR = Right Hippocampus; HippoL = Left Hippocampus; flWMR = frontal lobe WM right; flWML = frontal lobe WM left; plWMR = parietal lobe WM right; plWML = parietal lobe WM left; tlWMR = temporal lobe WM right; tlWML = temporal lobe WM left; ACgCR=Right ACgG  anterior cingulate gyrus; ACgCL=Left ACgG  anterior cingulate gyrus; EntR = Right Ent entorhinal area; EntL = Left Ent entorhinal area; MCgCR = Right MCgG  middle cingulate gyrus ;MCgCL = Left MCgG  middle cingulate gyrus; MFCR = Right MFC   medial frontal cortex; MFCL = Left MFC   medial frontal cortex; OpIFGR = Right OpIFG opercular part of the inferior frontal gyrus; OpIFGL = Left OpIFG opercular part of the inferior frontal gyrus; OrIFGR = Right OrIFG orbital part of the inferior frontal gyrus; OrIFGL = Left OrIFG orbital part of the inferior frontal gyrus; PCgCR = Right PCgG  posterior cingulate gyrus ; PCgCL = Left PCgG  posterior cingulate gyrus; PCuR = Right PCu   precuneus; PCuL = Left PCu   precuneus; SPLR  = Right SPL superior parietal lobule; SPLL  = Left SPL superior parietal lobule.}
\label{table:roi}
\end{table}
\end{document}